\documentclass[twocolumn,final]{svjour3}       
\smartqed  
\usepackage{amsmath}
\usepackage{amssymb}
\usepackage{nicefrac}

\usepackage{graphicx}
\usepackage{psfrag}

\usepackage{subfig}
\usepackage{bm}
\usepackage{enumitem}
\usepackage{color}

\usepackage{algorithm}
\usepackage[noend]{algpseudocode}

\usepackage[pagebackref=true,breaklinks=true,colorlinks,bookmarks=false,    linkcolor=blue,filecolor=blue,urlcolor=blue,citecolor=blue]{hyperref}

\usepackage[switch]{lineno}
\let\oldequation\equation
\let\oldendequation\endequation

\renewenvironment{equation}
  {\linenomathNonumbers\oldequation}
  {\oldendequation\endlinenomath}

\newcommand{\describeContent}[1]{%
\begingroup%
\let\thefootnote\relax%
\footnotetext{#1}%
\endgroup%
}

\journalname{Preprint: J. Math. Imaging Vision}
\begin{document}
\sloppy

\title{On the Generalized Essential Matrix Correction:}
\subtitle{An efficient solution to the problem and its applications}

\titlerunning{On the Generalized Essential Matrix Correction}        

\authorrunning{P. Miraldo and J. Cardoso} 

\author{Pedro Miraldo \and Jo\~ao R. Cardoso}

\institute{P. Miraldo \at
Institute for Systems and Robotics (ISR/IST), LARSyS, Instituto Superior Técnico, University of Lisboa, Portugal.
Part of this work was done when P. Miraldo was at the Division of Decision and Control Systems, KTH Royal Institute of Technology, Stockholm, Sweden.\\
\email{pedro.miraldo@tecnico.ulisboa.pt}
\and
J. R. Cardoso \at
Coimbra Polythecnic Institute (ISEC) and
Center for Mathematics, University of Coimbra,
Coimbra, Portugal.\\
\email{jocar@isec.pt}
}

\date{Received: \today}

\maketitle

\begin{abstract}
This paper addresses the problem of finding the closest generalized essential matrix from a given $6\times 6$ matrix, with respect to the Frobenius norm. To the best of our knowledge, this nonlinear constrained optimization problem has not been addressed in the literature yet. Although it can be solved directly, it involves a large number of constraints, and any optimization method to solve it would require much computational effort. We start by deriving a couple of unconstrained formulations of the problem. After that, we convert the original problem into a new one, involving only orthogonal constraints, and propose an efficient algorithm of steepest descent-type to find its solution. To test the algorithms, we evaluate the methods with synthetic data and conclude that the proposed steepest descent-type approach is much faster than the direct application of general optimization techniques to the original formulation with 33 constraints and to the unconstrained ones. To further motivate the relevance of our method, we apply it in two pose problems (relative and absolute) using synthetic and real data.
\end{abstract}

\section{Introduction}
The epipolar constraint is one of the fundamental geometry constraints in computer vision. It relates the rigid transformation between two cameras with different external parameters \cite{Ma,Hartley} and correspondences between points in the two images. It is one of the most common tools for scene reconstruction, known as passive techniques, i.e., two cameras looking at the same scene from different points of view. The epipolar constraint has been used in many other applications, such as visual odometry \cite{Nister}.

\begin{figure}
  \begin{center}
    \subfloat[Central Camera]{\includegraphics[width=0.22\textwidth,angle=0]{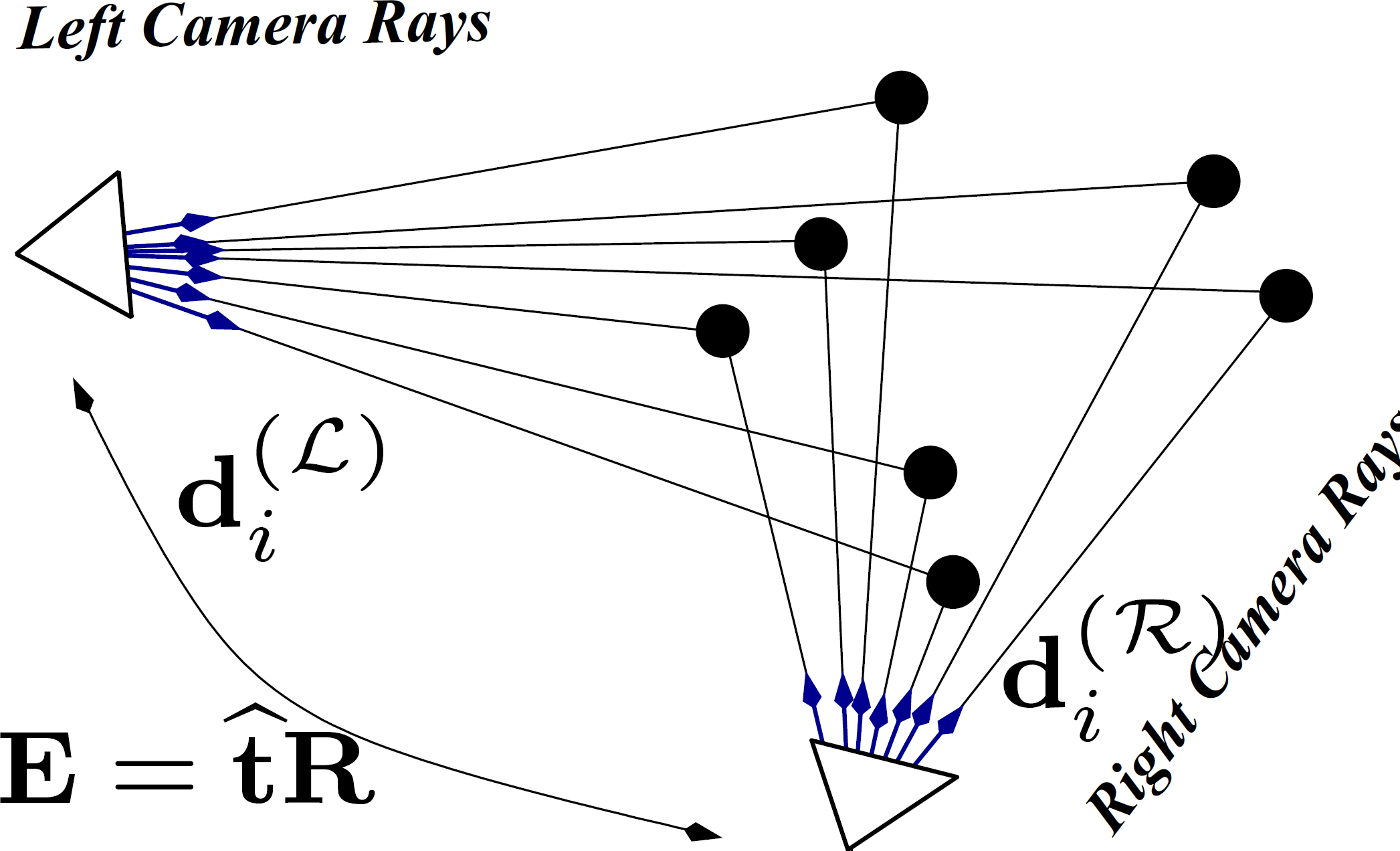}\label{fig:central_rep}}\,
    \subfloat[General Camera]{\includegraphics[width=0.23\textwidth,angle=0]{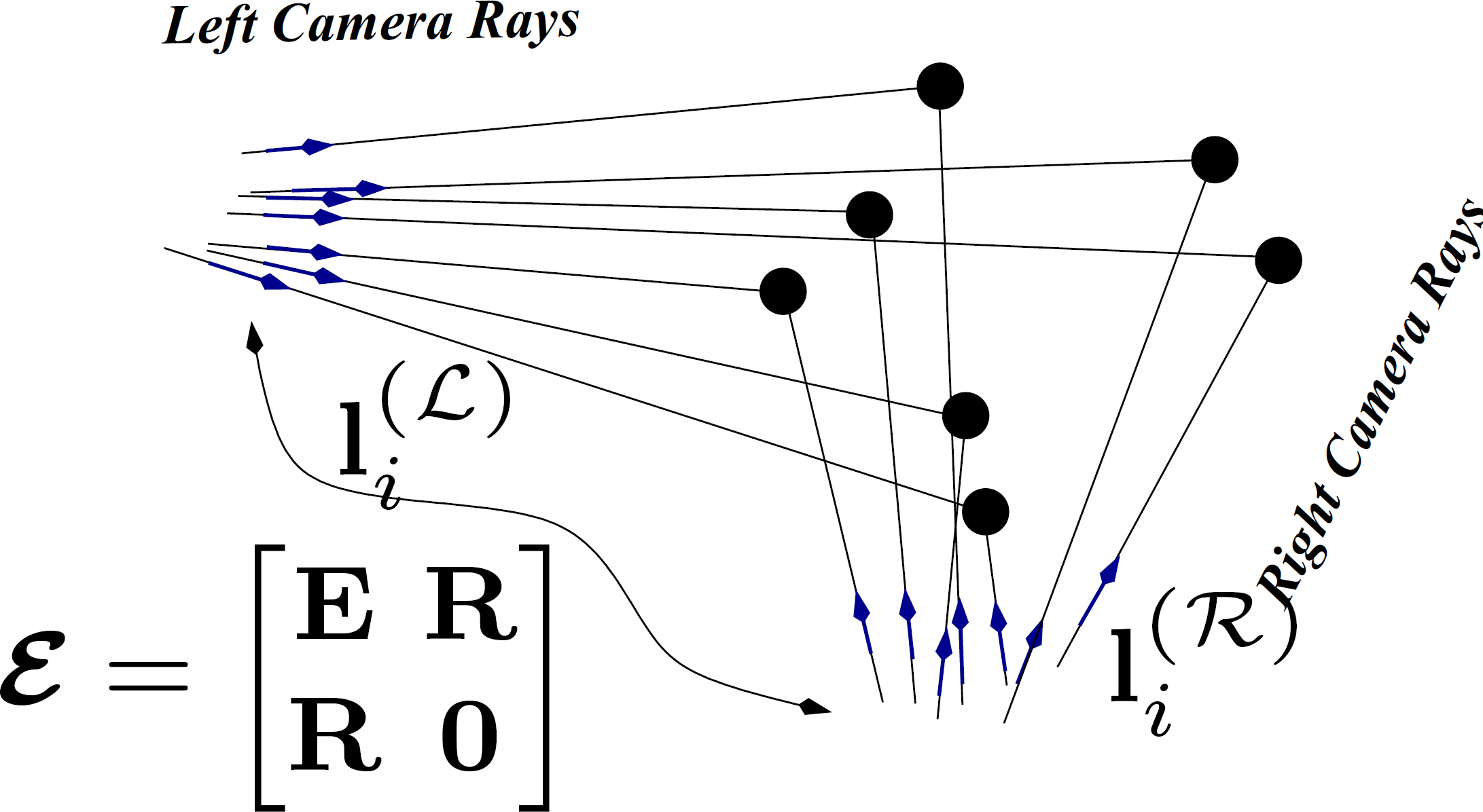}\label{fig:ncentral_rep}}
  \end{center}
  \caption{\it Central vs general camera systems, \protect\subref{fig:central_rep} and~\protect\subref{fig:ncentral_rep}, respectively. This paper addresses the general case, i.e., camera system does not have a single view point.}
  \label{fig:camera_sys}
\end{figure}

For many years, authors focused on perspective cameras to build this stereo pair \cite{Hartley}, see Fig.~\ref{fig:camera_sys}\subref{fig:central_rep}. However, these cameras have, among several disadvantages, a limited field of view. To overcome this, some authors have developed new camera systems. Special emphasis has been given to omnidirectional cameras, which get a larger field of view from a combination of perspective cameras with mirrors and/or fisheye lenses, the use of  \cite{Nalwa,Nayar,Micusik}, or multi-perspective camera systems \cite{Kim,Lee}. Most of these devices are non-central (see \cite{Swaminathan}). Other types of imaging sensors have been proposed. The perspective camera model cannot model most of them due to their physical constraints. Examples include pushbroom cameras \cite{Gupta} or refractive imaging devices \cite{Treibitz}. These kind of cameras are called non-central. To handle both central or non-central systems, we consider a pair of camera systems that are parameterized by the general camera model \cite{Grossberg,Sturm2,Miraldo3}. In this model, camera pixels are mapped into generic 3D straight lines in the world. If all these 3D lines intersect in a single point, the system is central (see Fig.~\ref{fig:camera_sys}\subref{fig:central_rep}), otherwise it is non-central (see Fig.~\ref{fig:camera_sys}\subref{fig:ncentral_rep}).

For a central camera stereo problem, one can define a $3\times 3$ matrix that encodes the epipolar constraint, i.e., the incident relation between the projection lines of both cameras~\cite{Ma,Hartley}. This matrix is called {\it essential}. For the case of general camera systems, such a matrix cannot be used because the central constraints are not satisfied. Instead, it can be proved that there is a $6\times 6$ matrix that expresses the incident relation between the projection lines of two general camera systems~\cite{Pless,Sturm,Miraldo}. Such a matrix is called {\it generalized essential} and has a particular block structure, involving rotation and skew-symmetric matrices of order $3$. Often, computer vision problems wherever this matrix has to be determined are affected with noise, and this may result in a matrix of order $6$ that fails to fit the structure characterizing the generalized essential matrices. In these cases, one needs to find the closest generalized essential matrix from a generic $6\times 6$ matrix. From a mathematical point of view, this is a nonlinear constrained matrix optimization problem. The fact that it is a nonconvex problem raises many difficulties to find a global minimum.

Examples of applications requiring the estimation of a generalized essential matrices are: 1) the computation of the minimal relative pose position for general camera models \cite{Stewenius}; 2) the estimation of the camera motion using multiple multi-perspective camera systems \cite{Kim,Lee}; 3) general structure-from-motion algorithms \cite{Ramalingam,Mouragnon}; and 4) the estimation of the camera absolute pose using known coordinates of 3D straight lines \cite{Miraldo2}. Since the particular structure of generalized essential matrices involves many nonlinear constraints (a rotation matrix and its product with a skew-symmetric matrix), finding the correct parameters of these matrices may slow down significantly the algorithms. An essential advantage of having a method to approximate general essential matrices from generic $6\times 6$ matrices is to allow us to get rid of some of those constraints (or, at least, to reduce the tolerance of these constraints in the optimization processes) to turn the algorithms faster.

In this paper, we show that the problem under investigation can be reformulated as an optimization problem with only orthogonal constraints. Then, we present an efficient algorithm to find a solution. Also, we give theoretical arguments to demonstrate why the problem has a global minimum (some open questions will be pointed out). We recall that methods for problems with orthogonal constraints are available in the literature \cite{Edelman,Jiang,Manton,Wen,Campos}, but their difficulties depend significantly on the objective function. In our problem, the objective function is not easy to handle, raising many challenging issues. To address them, we resort to techniques from matrix theory, and optimization on manifolds.

Suppose we have estimated an $6\times 6$ matrix $\mathbf{A}$ representing the epipolar geometry for general cameras, linearly obtained from the incident relation between of a set of inverse projection rays $\mathbf{l}_i^{(\mathcal{L})}$ and $\mathbf{l}_i^{(\mathcal{L})}$; i.e., building from the geometric constraints shown in Fig.~\ref{fig:camera_sys}\subref{fig:ncentral_rep} without imposing the structure of $\bm{\mathcal{E}}$.
The method proposed in this paper aims at approximating a generalized essential matrix $\bm{\mathcal{E}}$, from a general  $6\times 6$ matrix $\mathbf{A}$, with the assumption that the latter was estimated by ignoring some of the generalized essential constraints. Our motivation for developing such a method is twofold. When, for some reason, the estimation of $\mathbf{A}$ do not consider some of the generalized essential matrix constraints or ignore them altogether (such as DLT techniques), methods such as ours are helpful to obtain a real generalized essential matrix. When a large tolerance for the generalized essential constraints is used to speed up the computation of $\mathbf{A}$, methods like ours can be utilized to correct the result. Experiments illustrating the latter situation will be presented in Sec.~\ref{sec:real_applications_experiments}.

For the central case, we recall that similar approximation techniques have been used for pose estimation. In \cite{Hartley97}, the author estimates a real essential matrix with respect to the Frobenius norm by firstly using DLT techniques to compute a rough estimative to the $3\times 3$ essential matrix. He proved that this method performs almost as well as the best iterative algorithms, being faster in many cases. More recently, other methods have been developed using similar approximation techniques for the central perspective camera; see, for instance,  \cite{Yamaguchi,Zaragoza}. We believe that the method we are proposing in this paper will contribute to the development of similar techniques, but for general camera models. We recall that the goal of this work is not to propose a technique for the problem of estimating a generalized essential matrix from a set of projection lines (as did in \cite{Li} for general camera systems and \cite{Kneip} for multi-camera systems) but, instead, to estimate $\bm{\mathcal{E}}$ from a previously computed $\mathbf{A}$ by other techniques. The list of contributions of this paper are:
\begin{enumerate}
    \item It is the first work that investigates the fitting of general essential matrices from general $6\times 6$ matrices;
    \item We define the problem at hand and present a natural formulation of it by defining a non-linear constrained optimization problem;
    \item Based on the problem definition, we derive an unconstrained optimization version that can be used to solve the problem at hand;
    \item We change the initial formulation and write the problem as a function depending only on rotation unknowns;
    \item We use the formulation derived in the previous item to obtain an unconstrained optimization problem; and
    \item We propose an efficient method that iterates in the space of orthogonal matrices to obtain a solution to the problem. We prove that this method is the most efficient.
\end{enumerate}

\subsection{Outline of the Paper}
This paper is organized as follows. We start by revisiting the similar, but simpler, problem of finding the closest essential matrix arising in central camera systems. Then we formulate mathematically the problem under investigation in this work and explain how it could be solved straightforwardly. In Sec.~\ref{sec:efficient_solution}, we reformulate the original problem and derive two solutions for it. In Sec.~\ref{sec:experimental_results}, our method is compared with direct solutions using synthetic data. Results with the motivation on the use of our technique are shown in Sec.~\ref{sec:real_applications_experiments}. Experimental results are discussed in the respective experimental sections. Finally, in Sec.~\ref{sec:conclusions}, some conclusions are drawn.
\section{Notation}
\label{sec:notations}
Small regular letters denote scalars, e.g., $a$; small bold letters denote vectors, e.g., $\mathbf{a}\in\mathbb{R}^{n}$; and capital bold letters denote matrices, e.g., $\mathbf{A}\in\mathbb{R}^{n\times m}$. In a matrix $\mathbf{A}$, $\mathbf{A}_{(i:j,k:l)}\in\mathbb{R}^{(j-i+1)\times (l-k+1)}$ denotes the submatrix composed by the lines from $i$ to $j$ and columns from $k$ to $l$. We represent general projection lines using {\it Pl\"{u}cker} coordinates \cite{Pottmann}:
\begin{equation}
  \mathbf{l}\in\mathbb{R}^6 \sim \begin{bmatrix} \mathbf{d}^T & \mathbf{m}^T \end{bmatrix}^T \text{, such that}\;\; \mathbf{d}^T \mathbf{m} = 0
\end{equation}
where $\mathbf{d}\in\mathbb{R}^3$ and $\mathbf{m}\in\mathbb{R}^3$ are the line direction and moment, respectively. The operator $\sim$ denotes a vector that can be represented up to a scale factor.

The hat operator represents the skew-symmetric matrix that linearizes the cross product, i.e., $\mathbf{a} \times \mathbf{b} = \widehat{\mathbf{a}}\mathbf{b}$. $\left|\left| \mathbf{X} \right|\right|$ denotes the Frobenius norm (also known as $\mathcal{L}_2$-norm) of the matrix $\mathbf{X}$, which can be defined as a function of the trace:
\begin{equation}\label{eq_frobenius}
\left|\left| \mathbf{X} \right|\right|^2=\text{trace}\left(\mathbf{X}^T\mathbf{X}\right).
\end{equation}

$\mathcal{SO}(3)$ stands for the group of rotation matrices of order 3, i.e., the group of orthogonal matrices with determinant equal to 1.
To conclude, $\text{diag}(a_1,a_2,\dots,a_n)$ denotes an $n\times n$ diagonal matrix, with diagonal entries equals to $a_1,a_2,\dots,a_n$, $\text{expm}(\mathbf{A})$ represents the matrix exponencial of $\mathbf{A}$, and the asterisk symbol in $\mathbf{A}_\ast$ denotes a minimizer of an optimization problem.
\section{Problem Definition and Direct Solutions}
\label{sec:problem_def}

In this section, we start by defining the problem of fitting generalized essential matrices from general 6$\times$6 matrices and present two straightforward methods to solve it.

\subsection{Problem Definition}
In order to better understanding the generalized essential matrix parametrization, we first revisit the more straightforward case of the regular essential matrix corresponding to Fig.~\ref{fig:camera_sys}\subref{fig:central_rep}. An essential matrix aims at representing the incident relation between two projection lines of two cameras looking at the same 3D points in the world. The rigid transformation between both coordinate systems is taken into account.

Without loss of generality, we assume that both cameras are represented at the origin of each coordinate system, ensuring that all the 3D projection lines of each camera pass through the origin of the respective camera coordinate system. Under this assumption, one can represent 3D projection lines by the respective directions, here denoted as $\mathbf{d}^{(\mathcal{L})}_i$ and $\mathbf{d}^{(\mathcal{R})}_i$ for $i^{\text{th}}$ 3D projection lines that must intersect in the world, where $(\mathcal{L})$ and $(\mathcal{R})$ represent the left and right rays, respectively. Moreover, we assume that the transformation between both cameras is given by a rotation matrix $\mathbf{R}\in\mathcal{SO}(3)$ and a translation vector $\mathbf{t}\in\mathbb{R}^{3}$. Using this formulation, an essential matrix $\mathbf{E}\in\mathbb{R}^{3\times 3}$ is defined by:
\begin{equation}
  \left.\mathbf{d}_i^{(\mathcal{L})}\right.^T \mathbf{E}\mathbf{d}_i^{(\mathcal{R})} = 0, \; \text{such that} \;\; \mathbf{E} \doteq \widehat{\mathbf{t}}\mathbf{R}.
\end{equation}
Hence, the problem of finding the closest essential matrix $\mathbf{X}_\ast$ from a general $\mathbf{A}\in\mathbb{R}^{3\times 3}$ may be formulated as:
\begin{equation}
  \operatornamewithlimits{argmin}\limits_{\mathbf{X}}\, \| \mathbf{A} - \mathbf{X}\|\ , \mbox{s.t.}\ \mathbf{X}\in\mathbb{R}^{3\times 3}\ \mbox{is an essential matrix}.
\end{equation}
It has an explicit solution (see, for instance, Theorem~5.9 in \cite{Ma}):
\begin{equation}
  \mathbf{X}_\ast = \mathbf{U}\ \text{diag}(\sigma,\sigma,0)\ \mathbf{V}^T, \text{ with  } \sigma \doteq \frac{\lambda_1 + \lambda_2}{2}
\end{equation}
where the $3\times 3$ orthogonal matrices $\mathbf{U}$ and $\mathbf{V}$ and scalars $\lambda_1,\lambda_2$ are given by the singular value decomposition of $\mathbf{A}$:
\begin{equation}
  \mathbf{A} = \mathbf{U}\ \text{diag}(\lambda_1,\lambda_2,\lambda_3)\ \mathbf{V}.
\end{equation}

It turns out that, for the general case (the one addressed in this paper) and since the perspective constraints are note verified, we cannot represent the 3D projection lines only with its directions \cite{Grossberg}. One has to parametrize these 3D projection lines using a general 3D straight lines representation (unconstrained 3D straight lines), see Fig.~\ref{fig:camera_sys}\subref{fig:ncentral_rep}. Let us represent lines of both cameras as $\mathbf{l}^{(\mathcal{L})}_i$ and $\mathbf{l}^{(\mathcal{R})}_i$, represented in each coordinate system and parameterized by {\it Pl\"{u}cker} coordinates (see Sec.~\ref{sec:notations}). The incident relation between both sets of 3D projection lines is given by:
\begin{equation}
  \label{eq:generalizes_essential_matrix}
  \left.\mathbf{l}_i^{(\mathcal{L})}\right.^T \bm{\mathcal{E}}\ \mathbf{l}_i^{(\mathcal{R})} = 0, \; \text{such that} \;\; \bm{\mathcal{E}}\in\mathcal{X},
\end{equation}
where $\bm{\mathcal{E}}\in\mathbb{R}^{6 \times 6}$ is a {\it generalized essential} matrix \cite{Pless,Sturm,Miraldo} and $\mathcal{X}$ denotes the space of generalized essential matrices:
\begin{equation}
  \label{eq:generalized_essential_matrix_space_solutions}
  \mathcal{X}=\left\{\begin{bmatrix} \widehat{\mathbf{t}}\,\mathbf{R} & \mathbf{R} \\ \mathbf{R} & \mathbf{0} \end{bmatrix}:\ \mathbf{R} \in \mathcal{SO}(3), \widehat{\mathbf{t}}\ \mbox{is skew-symmetric}\right\}.
\end{equation}

As observed in \eqref{eq:generalized_essential_matrix_space_solutions}, a generalized essential matrix has a particular form. It is built up by block matrices that depend on rotation and translation parameters. Likewise the estimation of essential matrices, in many situations $\bm{\mathcal{E}}$ is estimated without enforcing the respective constraints. This happens, in particular, when using DLT techniques \cite{Hartley} or iterative schemes that do not take into account all the constraints, to speed up the respective process.

One of the goals of this paper is to estimate a real generalized essential matrix (i.e., a matrix satisfying the constraints associated with \eqref{eq:generalized_essential_matrix_space_solutions}) that is closest to a general $\mathbf{A}\in\mathbb{R}^{6\times 6}$ with respect to the Frobenius norm. Formally, this problem is formulated as:
\begin{equation}
  \begin{aligned}
    & \operatornamewithlimits{argmin}\limits_{\mathbf{X}}
    & & \left|\left|\mathbf{X}-\mathbf{A} \right|\right|^2 \\
    & \text{subject to}
    & & \mathbf{X} \in \mathcal{X}.
  \end{aligned}
  \label{eq:problem_def}
\end{equation}

Next, we present some naive approaches to solve \eqref{eq:problem_def}.

\subsection{A Direct Solution to the Problem in \eqref{eq:problem_def}}

The crucial part in solving \eqref{eq:problem_def} is to ensure that $\mathbf{X}$ belongs to the space of solutions $\mathcal{X}$. There is, however, a direct way of ensuring this, which can be derived directly from the constraints in \eqref{eq:generalized_essential_matrix_space_solutions}. These constraints are associated with the fact that $\mathbf{X}$ can be built up by stacking both $\mathbf{R}$ and $\widehat{\mathbf{t}}\mathbf{R}$, allowing us to rewrite the problem as:
\begin{equation}
  \begin{aligned}
    & \operatornamewithlimits{argmin}\limits_{\mathbf{X}}
    & & \left|\left|\mathbf{X}-\mathbf{A} \right|\right| \\
    & \text{subject to}
    & & \mathbf{X}_{(1:3,4:6)}^T \mathbf{X}_{(1:3,4:6)} = \mathbf{I} \\
    & & & \mathbf{X}_{(4:6,1:3)} - \mathbf{X}_{(1:3,4:6)} = \mathbf{0} \\
    & & & \mathbf{X}_{(4:6,4:6)} = \mathbf{0} \\
    & & & \mathbf{X}_{(1:3,1:3)} \mathbf{X}_{(1:3,4:6)}^T + \mathbf{X}_{(1:3,4:6)} \mathbf{X}_{(1:3,1:3)}^T  = \mathbf{0}.
    \end{aligned}
  \label{eq:trivial_prob_def}
\end{equation}

The main issues associated with the problem in \eqref{eq:trivial_prob_def} are related to the large amount of constraints. In total, 33 constraints are involved, being many of them quadratic. As we shall see in the experimental results, this will increase significantly the computational time required to fit the generalized essential matrix. To eliminate the high amount of constraints, in the next subsection, we reformulate the problem in \eqref{eq:problem_def} as an unconstrained one.

\subsection{Unconstrained Formulation of \eqref{eq:problem_def}}\label{sec:direct_general}

In this subsection, we derive an unconstrained formulation of the problem in \eqref{eq:problem_def}. Let $\mathcal{SE}(3)$ stand for the special Euclidean Lie group of motions in $\mathbb{R}^3$, that is, the group of matrices of the form
\begin{equation}
    \mathbf{T} = \begin{bmatrix}
    \mathbf{R} & \mathbf{t} \\ \mathbf{0} & 1
    \end{bmatrix}.
\end{equation}
The corresponding Lie algebra (i.e., the tangent space at the identity) is denoted by $\mathfrak{se}(3)$, which consists of matrices of the form
\begin{equation}
    \bm{\xi}(\bm{w},\bm{v}) = \begin{bmatrix}
    \widehat{\bm{w}} & \bm{v} \\ \mathbf{0} & 1
    \end{bmatrix},
    \label{eq:tangent_spaceT}
\end{equation}
where $\bm{w},\bm{v}\in\mathbb{R}^3$. It is well-known that the exponential map transforms 
a matrix in $\mathfrak{se}(3)$ to a matrix in 
$\mathcal{SE}(3)$:
\begin{equation}
    \text{expm}:\mathfrak{se}(3)\mapsto\mathcal{SE}(3),
    \label{eq:T_using_tangentspace}
\end{equation}
such that 
$\mathbf{T} = \text{expm}(\bm{\xi}(\bm{w},\bm{v})).$
Now, by inserting \eqref{eq:tangent_spaceT} and \eqref{eq:T_using_tangentspace} in \eqref{eq:problem_def}, we can rewrite the problem as:
\begin{equation}
  \begin{aligned}
    & \operatornamewithlimits{argmin}\limits_{(\bm{w},\bm{v})}
    & & \left|\left|\text{expm}(\bm{\xi}(\bm{w},\bm{v}))-\mathbf{A} \right|\right| \\
  \end{aligned},
  \label{eq:trivial_prob_def_Direc}
\end{equation}
thus resulting in a non-linear unconstrained problem with six unknowns. Although at first sight this problem might look easy to solve, we stress that the exponential map of a $4\times 4$ matrix may be computationally expensive; an efficient way of evaluating the map $\text{expm}:\mathfrak{se}(3)\mapsto \mathcal{SE}(3)$ in Sec.~\ref{sec:direct_general} is given by computing $\mathbf{R}$ as shown in \eqref{eq:efficient_exponential_map3}, where $\mu = \| \bm{w}  \|$ and $\mathbf{Z} = \widehat{\bm{w}}/\mu$, and $\mathbf{t}$ as:
\begin{equation}
  \mathbf{t} = \left(\mathbf{I} + \tfrac{1-\cos{(\mu_k)}}{\mu_k}\mathbf{Z}_k^{T} + \tfrac{(\mu_k-\sin{(\mu_k)})}{\mu_k}\left(\mathbf{Z}_k^{T}\right)^2\right)\bm{v}.
\end{equation}
In \cite{Cardoso10} another closed formulae to $\text{expm}(\bm{\xi}(\bm{w},\bm{v}))$ may be found. In addition, the fact that the we are not optimizing directly in $\mathbf{T}$ may lead to a high number of iterations and a less accurate solution.

In the next section, we derive an efficient algorithm for solving \eqref{eq:problem_def}, that exploits the particular features of the objective function and constraints.
\section{An Efficient Solution}
\label{sec:efficient_solution}

Now, we describe a method for an efficient approximation of the generalized essential matrix from a given $6\times 6$ matrix, with respect to the Frobenius norm. Our approach, first, represents the problem independently of the translation parameters, which means that only orthogonal constraints will be involved (Sec.~\ref{sec:reformulation_problem}).  Then, in Sec.~\ref{sec:solving_direct}, we use this new representation for deriving a new unconstrained optimization problem. Sec.~\ref{sec:solving_our_problem} proposes an efficient algorithm to solve the reformulated optimization problem, which is our main contribution.

\subsection{Reformulation of the Problem}\label{sec:reformulation_problem}
Since our goal is to represent the problem independently from the translation parameters, we aim at eliminating the skew-symmetric constraints in \eqref{eq:problem_def}. To ease the notation, we define the following submatrices: $\mathbf{A}_{11}\doteq \mathbf{A}_{(1:3,1:3)}$; $\mathbf{A}_{12}\doteq \mathbf{A}_{(1:3,4:6)}$; $\mathbf{A}_{21}\doteq \mathbf{A}_{(4:6,1:3)}$; and $\mathbf{A}_{22}\doteq \mathbf{A}_{(4:6,4:6)}$.

We start by finding a workable expression for the objective function in terms of $\mathbf{R}$ and $\widehat{\mathbf{t}}$:
\begin{eqnarray}
  f(\mathbf{R},\mathbf{t})&\doteq&\left\|\mathbf{X}-\mathbf{A}\right\|^2  \nonumber \\
&=&\left\| \begin{bmatrix}
           \mathbf{A}_{11} & \mathbf{A}_{12} \\
           \mathbf{A}_{21} & \mathbf{A}_{22}
    \end{bmatrix}
    -
    \begin{bmatrix}
           \widehat{\mathbf{t}}\mathbf{R} & \mathbf{R} \\
           \mathbf{R} & \mathbf{0}
    \end{bmatrix}
    \right\|^2 \nonumber \\
&=& \| \mathbf{A}_{11}-\widehat{\mathbf{t}}\mathbf{R}\|^2+\| \mathbf{A}_{12}-\mathbf{R}\|^2+ \nonumber\\
&& \| \mathbf{A}_{21}-\mathbf{R}\|^2+\| \mathbf{A}_{22}\|^2   \nonumber\\
&=& \| \mathbf{A}_{11}-\widehat{\mathbf{t}}\mathbf{R}\|^2+\nonumber\\
&& \text{trace}\left( (\mathbf{A}^T_{12}-\mathbf{R}^T)(\mathbf{A}_{12}-\mathbf{R}) \right)+ \nonumber\\
&& \text{trace}\left( (\mathbf{A}^T_{21}-\mathbf{R}^T)(\mathbf{A}_{21}-\mathbf{R}) \right)+\| \mathbf{A}_{22}\|^2 .\ \ \ \ \ \
\end{eqnarray}

Attending to the linearity of the $\text{trace}$ function and the fact that $\|\mathbf{R}^T\mathbf{R}\|^2=\|\mathbf{I}\|^2=3$, the following expression is obtained for the objective function:
\begin{multline}\label{eq:objective1}
f(\mathbf{R},\mathbf{t})=\|\mathbf{A}_{11}-\widehat{\mathbf{t}}\mathbf{R}\|^2-\\ 2\ \text{trace}\left((\mathbf{A}_{12}+\mathbf{A}_{21})^T\mathbf{R}\right)+\alpha,
\end{multline}
where:
\begin{equation}
  \alpha\doteq 6+\left\|\begin{bmatrix} \mathbf{0} & \mathbf{A}_{12} \\ \mathbf{A}_{21} & \mathbf{A}_{22}\end{bmatrix}\right\|^2
\end{equation}
is a constant. Let us denote $\mathbf{M}\doteq \mathbf{A}_{11}$ and $\mathbf{N}\doteq (\mathbf{A}_{12}+\mathbf{A}_{21})^T$. With respect to the Frobenius norm, it is well-known and easy to show that the nearest skew-symmetric $\widehat{\mathbf{t}}_\ast$ from a given generic matrix $\mathbf{B}\in\mathbb{R}^{3\times 3}$ is the so-called skew-symmetric part of $\mathbf{B}$ (check Theorem 5.3 in \cite{Cardoso} with $\mathbf{P}=\mathbf{I}$):
\begin{equation}
  \widehat{\mathbf{t}}_\ast=\frac{\mathbf{B}-\mathbf{B}^T}{2}.
\end{equation}
Hence, we can replace $\widehat{\mathbf{t}} \doteq \nicefrac{\left(\mathbf{M}\mathbf{R}^T-(\mathbf{M}\mathbf{R}^T)^T\right)}{2}$ in \eqref{eq:objective1}, yielding:
\begin{equation}
  g(\mathbf{R})\doteq\frac{1}{4}\ \left\|\mathbf{M}+\mathbf{R}\mathbf{M}^T\mathbf{R}\right\|^2-2\text{trace}(\mathbf{N}\mathbf{R})+\beta,
\end{equation}
where:
\begin{equation}\label{eq_beta}
\beta\doteq 6+\left\|\begin{bmatrix}\mathbf{0} & \mathbf{A}_{12} \\ \mathbf{A}_{21} & \mathbf{A}_{22} \end{bmatrix} \right\|^2+\frac{1}{2}\|\mathbf{A}_{11}\|^2.
\end{equation}
Writing again the Frobenius norm in terms of the trace of a matrix gives:
\begin{equation}\label{eq:objective2}
g(\mathbf{R})=\frac{1}{2}\ \text{trace}\left((\mathbf{M}^T\mathbf{R})^2\right)-
2\ \text{trace}\left(\mathbf{N}\mathbf{R}\right)+\beta,
\end{equation}
with $\beta$ being the constant given in \eqref{eq_beta}.
This allows a new reformulation of the problem \eqref{eq:problem_def} as:
\begin{equation}
  \begin{aligned}
    & \operatornamewithlimits{argmin}\limits_{\mathbf{R}}
    & & g(\mathbf{R}) \\
    & \text{subject to} & & \mathbf{R} \in \mathcal{SO}(3),
  \end{aligned}\label{sec:problem_reform}
\end{equation}
which has only orthogonal constraints.

Before we present our efficient solution to the problem at hand, we propose an unconstrained version of the problem in \eqref{sec:problem_reform}.

\subsection{An Unconstrained Formulation of \eqref{sec:problem_reform}}\label{sec:solving_direct}

We proceed similarly as in Sec.~\ref{sec:direct_general}. Let $\bm{w}\in \mathbb{R}^3$ and consider its corresponding skew-symmetric matrix $\widehat{\bm{w}}\in \mathfrak{so}(3)$. Using the {\it Lie} mapping, one can write
\begin{equation}
    \text{expm}:\mathfrak{so}(3)\mapsto\mathcal{SO}(3),
    \ \text{such that} \ \ \mathbf{R} = \text{expm}(\widehat{\bm{w}}).
    \label{eq:R_using_tangentspace}
\end{equation}

Now, by using \eqref{eq:R_using_tangentspace}, the problem in \eqref{sec:problem_reform} can be rewritten as:
\begin{equation}
  \begin{aligned}
    & \operatornamewithlimits{argmin}\limits_{\bm{w}}
    & & g(\text{expm}(\widehat{\bm{w}})),
  \end{aligned}\label{sec:problem_reform_naive}
\end{equation}which is unconstrained. Efficient techniques to compute the exponential map  in~\eqref{eq:R_using_tangentspace} are available (see \eqref{eq:efficient_exponential_map3}). However, the fact that we are not iterating directly in the space of rotation matrices will bring inconvenient issues such as the requirement of a large number iterations.

In the next subsection, we aim at tackling these issues by proposing an efficient solution to the problem~\eqref{sec:problem_reform}.

\subsection{An Efficient Solution to \eqref{sec:problem_reform}}\label{sec:solving_our_problem}
Many optimization problems with orthogonal constraints have been investigated in the last two decades; see, for instance, \cite{Edelman,Manton,Wen,Abrudan,Campos}. The right framework to deal with this kind of problems is to regard them as optimization problems on matrix manifolds. Tools from Riemannian geometry, calculus on matrix manifolds, and numerical linear algebra are required. Similar techniques can be used in our particular problem \eqref{sec:problem_reform}, because the set of rotation matrices $\mathcal{SO}(3)$ is a manifold. It also has a structure of Lie group (see \cite{Absil}) and is a compact set. We recall that this latter property guarantees the existence of at least a global minimum for \eqref{sec:problem_reform}; see Part III of \cite{Luenberger}. However, the complicated expression of the objective function $g(\mathbf{R})$ turns hard to find an explicit expression for those global minima. Besides, the lack of convexity of our problem (neither the objective function nor the constraints are convex) may only guarantee the approximation of local minima.

It turns out, however, that some numerical experiments (not showed in this paper) suggest that the approximation produced by Algorithm~\ref{alg:ours} is indeed the global one. We could observe this because different initial guesses $\mathbf{X}_0$ led to convergence to the same matrix. Unfortunately, a theoretical confirmation that Algorithm~\ref{alg:ours} always converges to a global minimum is still unknown. Nevertheless, it can be guaranteed that when $\mathbf{A}$ is close to being a generalized essential matrix (this happens in many practical situations, as shown later in Section \ref{sec:experimental_results}), a local minimizer for \eqref{sec:problem_reform} will be very close to the global one. Now, we provide more insight into this claim. Let us consider the original formulation \eqref{eq:problem_def}, $\widetilde{\mathbf{X}}$ a local minimizer, and $\mathbf{X}_\ast$ a global minimizer. Assume that:
\begin{equation}
  \|\mathbf{A}-\widetilde{\mathbf{X}}\|=\epsilon,
\end{equation}
for a certain positive value $\epsilon$. Since $\|\mathbf{A}-\mathbf{X}_\ast\| \leq \epsilon$, we have:
\begin{eqnarray}
  \|\widetilde{\mathbf{X}}-\mathbf{X}_\ast\|&=&\|\tilde{\mathbf{X}}-\mathbf{A}+\mathbf{A}-\mathbf{X}_\ast\| \nonumber \\
  &\leq &\|\widetilde{\mathbf{X}}-\mathbf{A}\|+\|\mathbf{A}-\mathbf{X}_\ast\| \nonumber \\
 &\leq & 2\epsilon,
 \end{eqnarray}
 which shows that:
 \begin{equation}
   0\leq \|\widetilde{\mathbf{X}}-\mathbf{X}_\ast\| \leq 2\epsilon.
\end{equation}
This means that if $\epsilon\approx 0$ then $\widetilde{\mathbf{X}}\approx \mathbf{X}_\ast$. For instance, if $\epsilon=10^{-1}$, then $\widetilde{\mathbf{X}}\approx \mathbf{X}_\ast$ with an error not greater that $2\times 10^{-1}$. Note that both $\widetilde{\mathbf{X}}$ and  $\mathbf{X}_\ast$ are generalized essential matrices.

Once a local minimum $\mathbf{R}_\ast$ of \eqref{sec:problem_reform} has been computed, the corresponding skew-symmetric matrix $\widehat{\mathbf{t}}_\ast$ needed in \eqref{eq:problem_def} will be:
\begin{equation}
  \widehat{\mathbf{t}}_\ast=\frac{\mathbf{M}\mathbf{R}_\ast^T-\left(\mathbf{M}\mathbf{R}_\ast^T\right)^T}{2}.
\end{equation}
Hence, the required nearest generalized essential matrix from $\mathbf{A}$ will be given by:
\begin{equation}
  \mathbf{X}_\ast = \begin{bmatrix}\widehat{\mathbf{t}}_\ast\mathbf{R}_\ast & \mathbf{R}_\ast \\ \mathbf{R}_\ast & \mathbf{0} \end{bmatrix}.
\end{equation}

The algorithm we will propose is of the steepest-descent type. Loosely speaking, these methods are essentially based on the property that the negative of the gradient of the objective function points out the direction of fastest decrease. For more details on general steepest descent methods, see \cite[Sec. 8.6]{Luenberger} or \cite[Ch. 3]{Nocedal}. In our situation, one has to account the constraint that $\mathbf{R}$ must be a rotation matrix and so our algorithm will evolve on the manifold $\mathcal{SO}(3)$. Hence we have to resort to steepest descent methods on matrix manifolds
(see \cite[Ch. 3]{Absil}).

For particular manifolds, tailored methods exploiting its unique features have been proposed by many authors. For instance, for the complex Stiefel manifold, Manton \cite[Alg. 15]{Manton} proposed a modified steepest descent method based on Euclidean projections, where the Armijo's step-size rule calculates the length of the descent direction. This method has been adapted and improved by Abrudan et al. in \cite[Table II]{Abrudan} for the manifold of unitary matrices, where geodesics replace the Euclidean projections. When dealing with manifolds that are Lie groups, geodesics are defined upon the matrix exponential, which is a much-studied matrix function \cite{Moler}.

The strategy adopted for the steepest descent algorithm to be described below has been inspired in \cite{Manton,Abrudan}, and we refer the reader to those papers for more technical details. The particular nature of the objective function (\ref{eq:objective2}) is exploited in order to improve the efficiency of the method. In particular, we propose a specific technique to choose an initial guess (thus reducing the number of iterations) and avoid expensive methods based on Schur decompositions and Pad\'e approximation for the computation of matrix exponentials.

Before displaying the steps of our algorithm, one needs to find the Euclidean and the Riemannian gradients of the objective function. After some calculations (see \cite{Lutkepohl} for formulae on derivatives of the $\text{trace}$ function), the Euclidean gradient of $g$ at \eqref{eq:objective2} is:
\begin{equation}
  \nabla g(\mathbf{X})\doteq\mathbf{M}\mathbf{X}^T\mathbf{M}-2\mathbf{N}^T,
\end{equation}
and the Riemaniann gradient is:
\begin{equation}
  \text{grad}\, g(\mathbf{X})\doteq\nabla g(\mathbf{X})-\mathbf{X}\nabla g(\mathbf{X})^T\mathbf{X}.
\end{equation}

Note that $\text{grad}\, g(\mathbf{X})\mathbf{X}^T$ is a ``tangent vector'' that is actually a skew-symmetric matrix. Geodesics on $\mathcal{SO}(3)$ (i.e., curves giving the shortest path between two points in the manifold) can be defined through the matrix exponential as:
\begin{equation}
  G(t)=G(0)\,\ \text{expm}(\mu \mathbf{S}),
\end{equation}
where $\mathbf{S}\in\mathbb{R}^{3\times 3}$ is a skew-symmetric matrix representing a translation and $\mu$ is a real scalar. Algorithm \ref{alg:ours} summarizes the main steps of our method.

\vspace{.25cm}
\noindent
{\bf Algorithm:}
In a few words, Algorithm~\ref{alg:ours} starts with an initial approximation $\mathbf{X}_0\in \mathcal{SO}(3)$, finds the skew-symmetric matrix $\left(\text{grad}\, g(\mathbf{X})\right)\mathbf{X}^T$ (the gradient direction on the manifold), and performs several steps along geodesics until convergence. The positive scalar $\mu_k$ controls the length of the ``tangent vector'' and, in turn, the overall convergence of the algorithm. To find an almost optimal $\mu_k$, the algorithm uses the Armijo's step-size rule, as performed in \cite{Manton}.

\begin{algorithm}[t]
  \caption{\it Given a $6\times 6$ matrix $\mathbf{A}\in\mathbb{R}^{6\times 6}$, this algorithm approximates the closest {\it generalized essential} matrix $\bm{\mathcal{E}}\in\mathcal{X}$ from $\mathbf{A}$ for a given tolerance {\tt tol}.}
  \label{alg:ours}
  \begin{algorithmic}[1]
    \State $\mathbf{M} \gets \mathbf{A}_{11}$;
    \State $\mathbf{N} \gets (\mathbf{A}_{12}+\mathbf{A}_{21})^T$;
    \State $k \gets0$;
    \State $\mathbf{X}_0\in \mathcal{SO}(3)$ is an initial guess;
    \State $\mu_0 \gets 1$;
    \State {\tt error} $\gets 1$;
    \State Choose a tolerance {\tt tol};
    \While {\tt error > tol}
    \State $\nabla g(\mathbf{X}_k) \gets \mathbf{M}\mathbf{X}_k^T\mathbf{M}-2\mathbf{N}^T$;
    \State $\mathbf{Z}_k\gets\nabla g(\mathbf{X}_k)\, \mathbf{X}_k^T-\mathbf{X}_k\nabla g(\mathbf{X}_k)^T$;
    \State $z_k\gets0.5\ \text{trace}(\mathbf{Z}_k\mathbf{Z}_k^T)$;
    \State $\mathbf{P}_k\gets\ \text{expm}(-\mu_k\mathbf{Z}_k)$;
    \State $\mathbf{Q}_k\gets\mathbf{P}_k^2$;
    \While{$g(\mathbf{X}_k)-g(\mathbf{Q}_k\mathbf{X}_k)\geq \mu_kz_k$}
    \State $\mathbf{P}_k\gets\mathbf{Q}_k$;
    \State $\mathbf{Q}_k\gets\mathbf{P}_k\mathbf{P}_k$;
    \State $\mu_k\gets2\mu_k$;
    \EndWhile
    \While{$g(\mathbf{X}_k)-g(\mathbf{Q}_k\mathbf{X}_k) < 0.5\mu_kz_k$}
    \State $\mathbf{P}_k\gets\ \text{expm}(-\mu_k\mathbf{Z}_k)$;
    \State $\mu_k\gets0.5 \mu_k$;
    \EndWhile
    \State $\mathbf{X}_{k+1}\gets\mathbf{P}_k\mathbf{X}_k$;
    \State {\tt error}$\ \gets \|X_{k+1}-X_k\|$;
    \State $k\gets k+1$;
    \EndWhile \label{euclidendwhile}
    \State $\mathbf{R}\gets\mathbf{X}_{k}$;
    \State $\widehat{\mathbf{t}}\ =\ 0.5\ (\mathbf{M}\mathbf{R}^T-\mathbf{R}\mathbf{M}^T)$;
    \State $\bm{\mathcal{E}}\gets \begin{bmatrix} \widehat{\mathbf{t}}\mathbf{R} & \mathbf{R} \\ \mathbf{R} & \mathbf{0} \end{bmatrix}$.
  \end{algorithmic}
\end{algorithm}

\vspace{.25cm}
\noindent
{\bf Computational Remarks:}
Now, we draw our attention to some essential computational remarks about the algorithm.
\begin{itemize}
  \item{As the algorithm runs, the function $g$, which involves the computation of traces of products of matrices, is called several times. Note that the efficient computation of $\ \text{trace}(\mathbf{A}\mathbf{B})$ does not require matrix products. Instead, it can be carried out through the formula:
      \begin{equation}
        \label{eq:trace_simplification}
        \text{trace}(\mathbf{A}\mathbf{B})= \sum_{i,j}{(\mathbf{A}\circ \mathbf{B}^T)}_{(i,j)}, 
      \end{equation}
    where the operator $\circ$ denotes the Hadamard product, i.e., entry-wise product. If $\mathbf{A}$ and $\mathbf{B}$ are matrices of order $n$, the direct computation of the matrix product $\mathbf{A}\mathbf{B}$ needs $O(n^3)$ operations, while the trace at \eqref{eq:trace_simplification} just requires $O(n^2)$;}
  \item{The exponential of the $3\times 3$ skew-symmetric matrix $-\mu_k \mathbf{Z}_k$ in lines 12 and 19 of the algorithm can computed by means of the well-known Rodrigues' formula  \cite{Murray}:
      \begin{multline} 
        \label{eq:efficient_exponential_map3}
        \text{expm}(-\mu_k\mathbf{Z}_k) = \\ \mathbf{I} + \sin{(\mu_k)}\mathbf{Z}_k^{T} + (1-\cos{(\mu_k)})\left(\mathbf{Z}_k^{T}\right)^2,
      \end{multline}
      which involves (at leading cost) the computation of just one matrix product.
      The direct use of conventional $\texttt{expm}(.)$ functions based on the scaling and squaring method combined with Pad\'e approximation would be much more expensive. 
    \item{Note that the trace of any skew-symmetric matrix $\mathbf{S}$ is always zero and so:
        \begin{equation}\det(\text{expm}(\mathbf{S}))=\text{expm}(\text{trace}(\mathbf{S}))=1.
        \end{equation}
      This guarantees that matrices $\mathbf{X}_k$ do not leave the rotation manifold $\mathcal{SO}(3)$; and}
    \item{To conclude, we remark that the choice of the initial guess $\mathbf{X}_0$ influences the running time of the algorithm. An obvious choice for $\mathbf{X}_0$ would be the identity matrix $\mathbf{I}$ of order $3$. It turns out that other choices of $\mathbf{X}_0$ may reduce the number of iterations significantly in the algorithm. In our experiments, we have chosen as the initial guess the rotation matrix that maximizes  $2\text{trace}\left(\mathbf{N}\mathbf{R}\right)$ of the sum defining $g(\mathbf{R})$ in \eqref{eq:objective2}. We recall that this problem has an explicit solution based on the singular value decomposition of $\mathbf{N}$ (see \cite[Sec. 12.4]{Golub}):
        \begin{equation}
          \widetilde{\mathbf{R}}=\mathbf{U}\,\text{diag}\left(1,1,\text{det}(\mathbf{UV}^T)\right)\,\mathbf{V}^T,
    \end{equation}
    with $\mathbf{U}$ and $\mathbf{V}$ being the orthogonal matrices arising in the singular value decomposition of $\mathbf{N}^T$, that is, $\mathbf{N}^T=\mathbf{UDV}^T$.
    Since, in general, $g(\widetilde{\mathbf{R}})\leq g(\mathbf{I})$, it is expected that $\mathbf{X}_0=\widetilde{\mathbf{R}}$ will be more close to the minimizer than $\mathbf{X}_0=\mathbf{I}$.}}
\end{itemize}
\section{Implementation of Our Method}
\label{sec:experimental_results}

\begin{figure*}[t]
  \subfloat[\it Computational speed, {\tt Tolerance} of $10^{-9}$.]{
  \includegraphics[height=0.14\textheight]{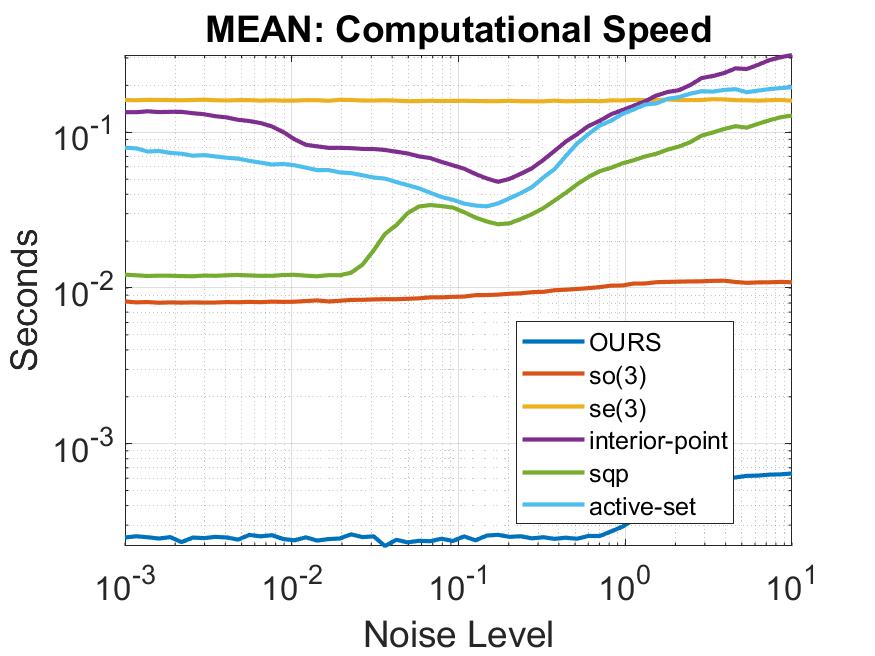}\nolinebreak
  \includegraphics[height=0.14\textheight]{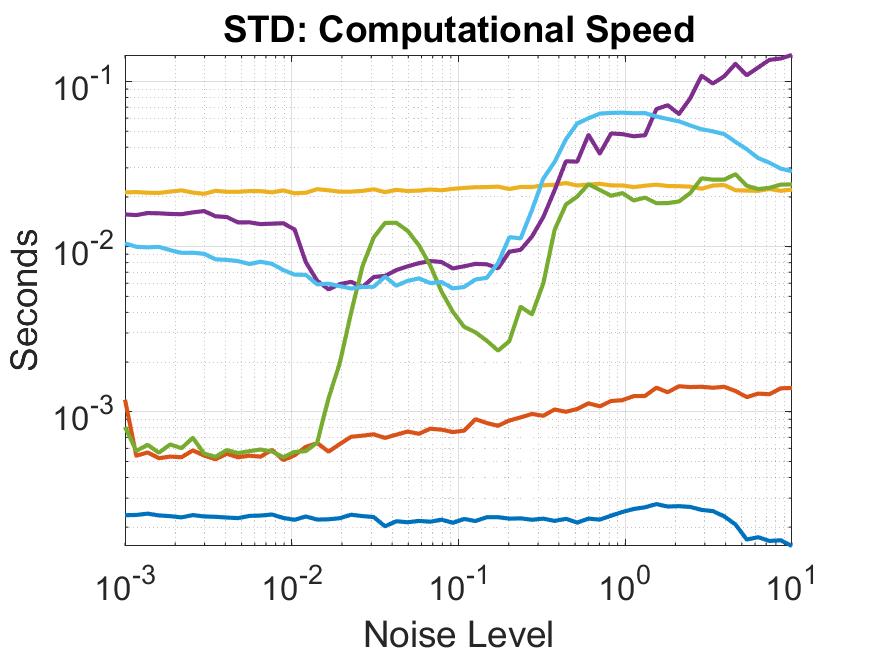}\label{fig:results_evaluation_noise_speed_e9}}
  \subfloat[\it Number of iterations, {\tt Tolerance} of $10^{-9}$.]{
  \includegraphics[height=0.14\textheight]{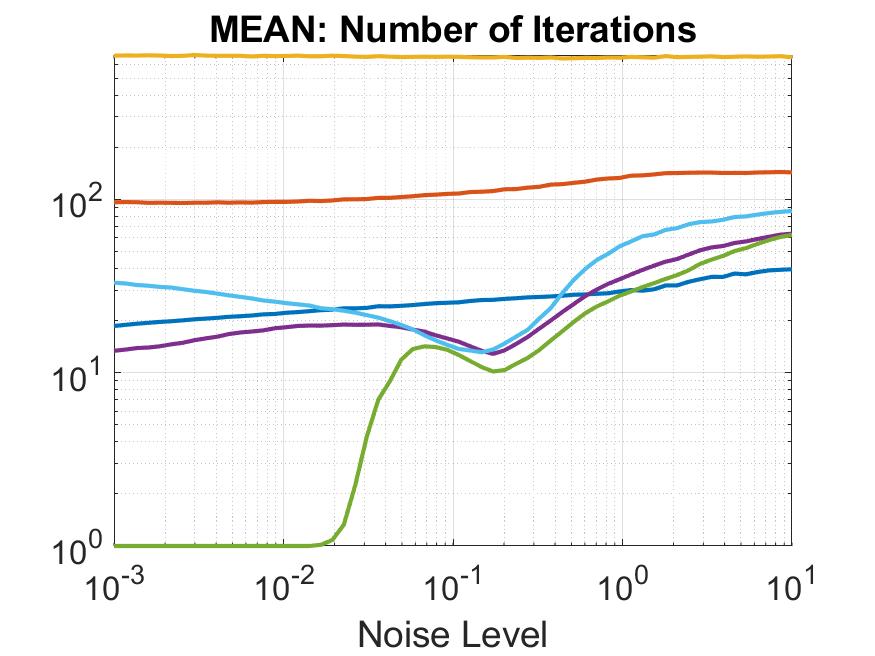}\nolinebreak
  \includegraphics[height=0.14\textheight]{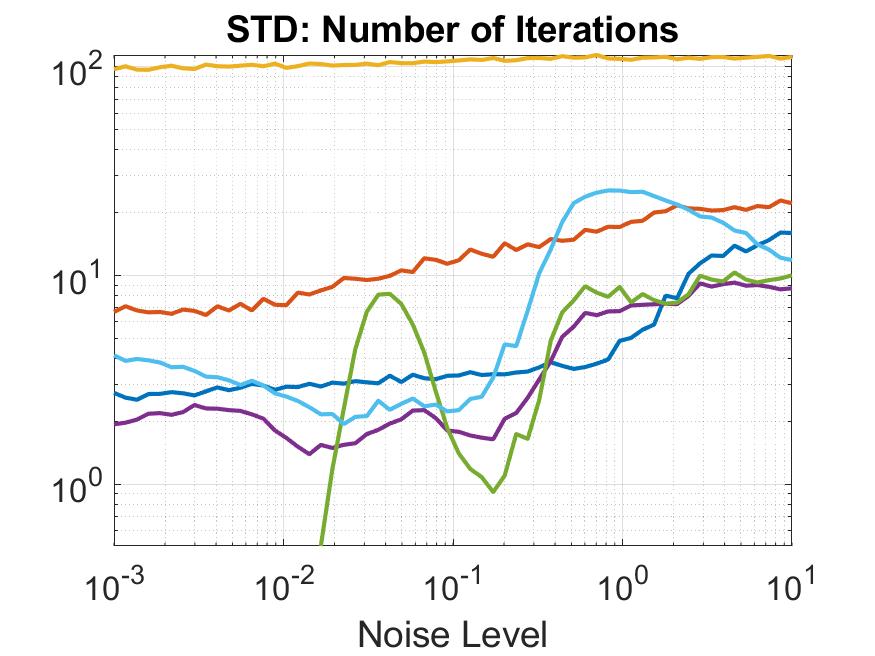}\label{fig:results_evaluation_noise_iter_e9}} \\
  \subfloat[\it Computational speed, {\tt Tolerance} of $10^{-6}$.]{
  \includegraphics[height=0.14\textheight]{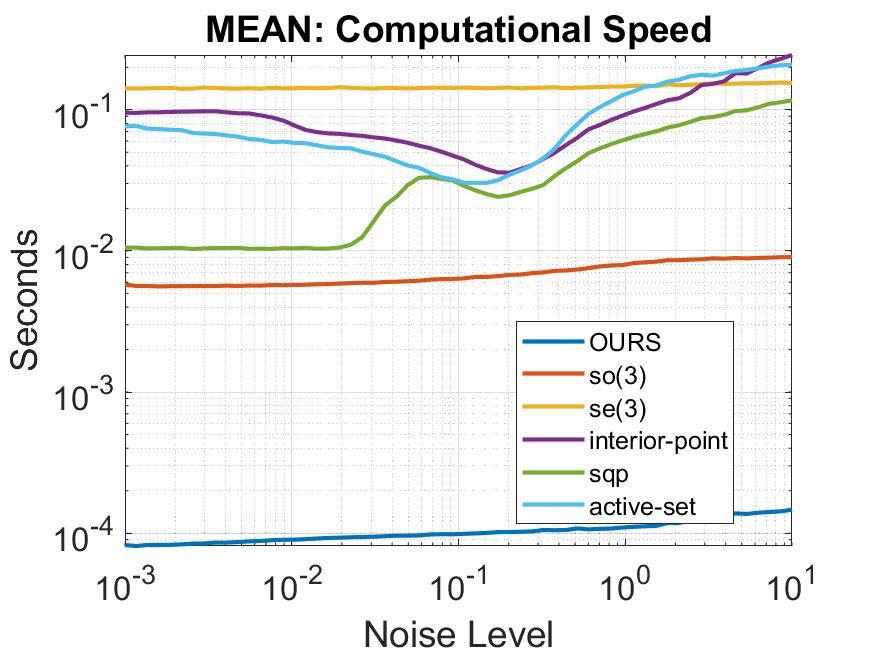}\nolinebreak
  \includegraphics[height=0.14\textheight]{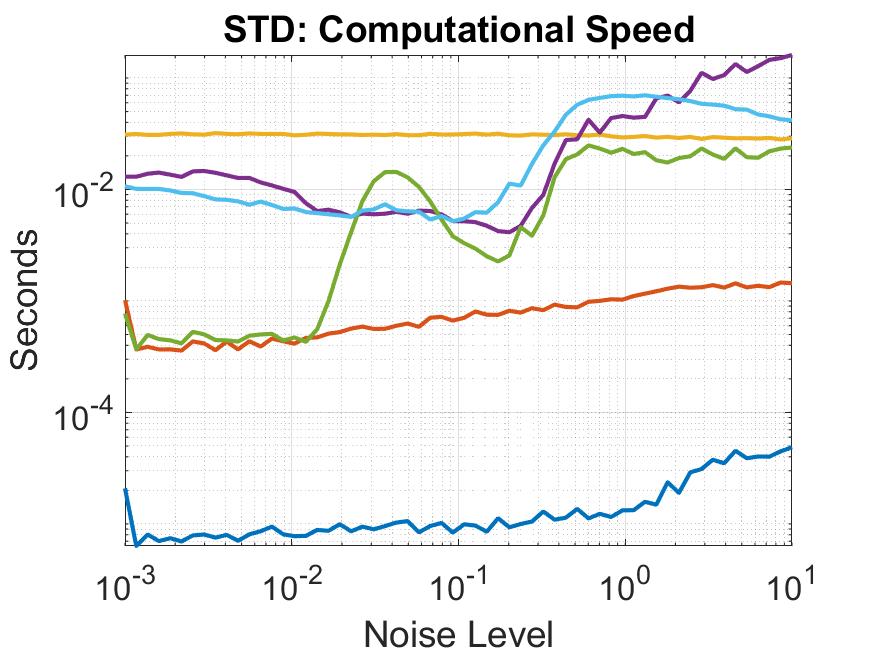}\label{fig:results_evaluation_noise_speed_e6}}
  \subfloat[\it Number of iterations, , {\tt Tolerance} of $10^{-6}$.]{
  \includegraphics[height=0.14\textheight]{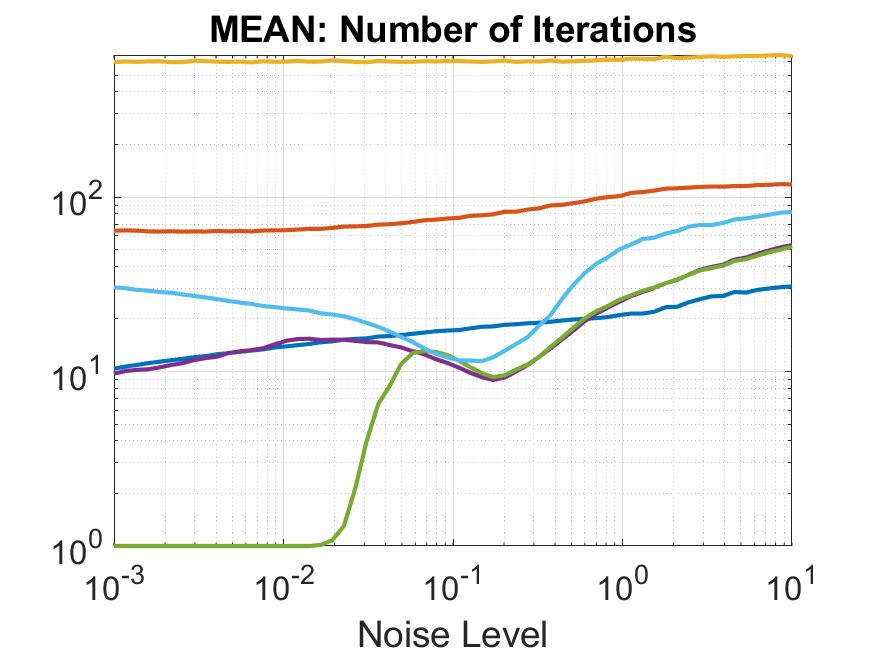}\nolinebreak
  \includegraphics[height=0.14\textheight]{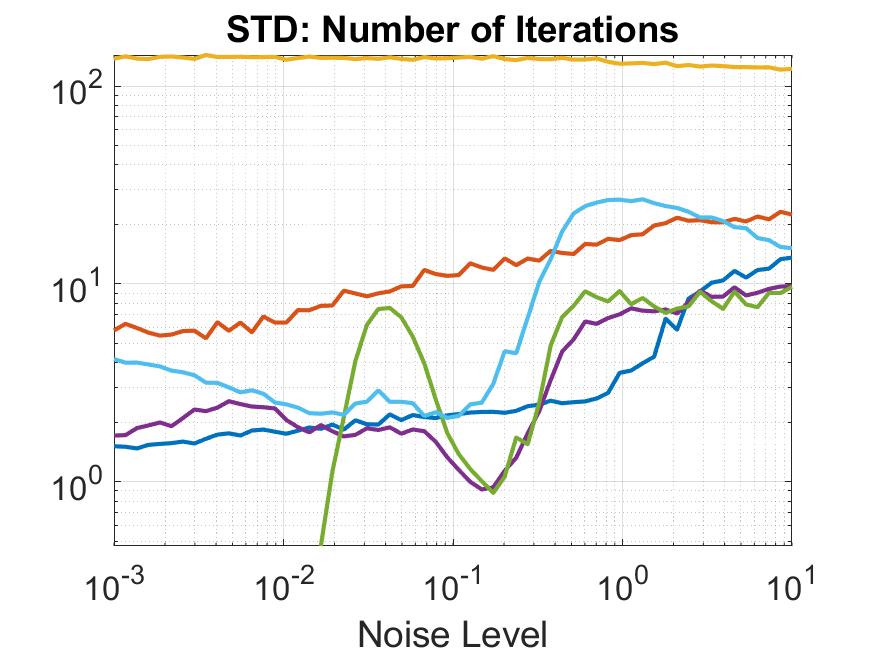}\label{fig:results_evaluation_noise_iter_e6}}
  \caption{\it Comparison between the method presented in this paper against general optimization techniques applied to \eqref{eq:trivial_prob_def}, as a function of the variation of the {\tt Noise Level}. The {\tt Tolerance} value is set to $10^{-9}$ and $10^{-6}$. We evaluate the methods for both the computational speed and the number of iterations, Figs.~\protect\subref{fig:results_evaluation_noise_speed_e9} \& \protect\subref{fig:results_evaluation_noise_iter_e9}, and \protect\subref{fig:results_evaluation_noise_speed_e6} \& \protect\subref{fig:results_evaluation_noise_iter_e6}, for the {\tt Tolerance} level of $10^{-9}$ and $10^{-6}$, respectively.}
  \label{fig:results_evaluation_noise}
\end{figure*}

In this section, we compare Algorithm~\ref{alg:ours} with general optimization techniques applied to the direct formulation of the problem \eqref{eq:trivial_prob_def} and the methods presented in Secs.~\ref{sec:direct_general} and \ref{sec:solving_direct}, using synthetic data.
We run all these experiments a single thread {\tt Intel i7-5820K} processor at 3.30GHz and 16GB of RAM.

\begin{figure*}[t]
  \subfloat[\it Computational speed, {\tt Noise Level} of $10^{-1}$.]{
  \includegraphics[height=0.14\textheight]{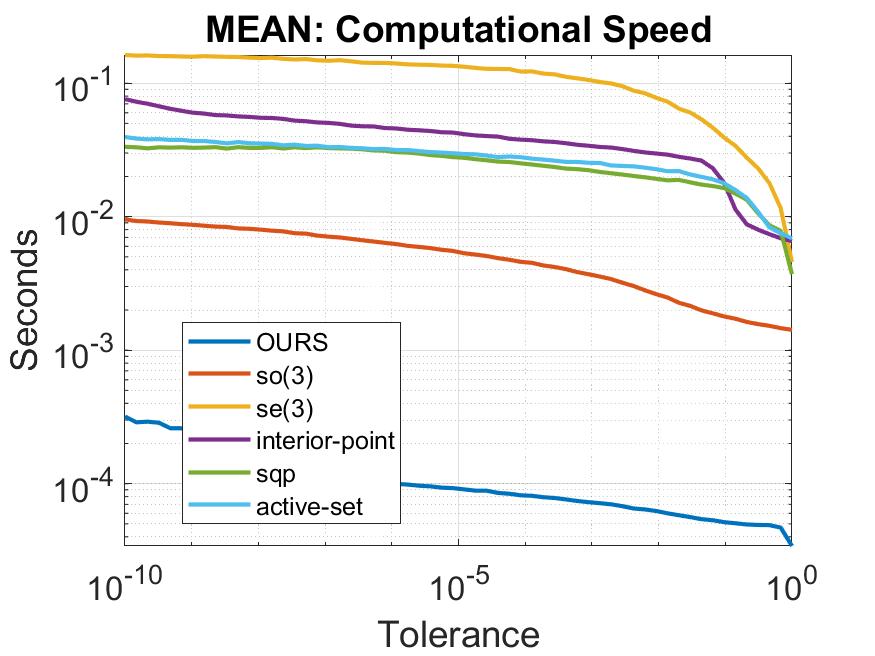}\nolinebreak
  \includegraphics[height=0.14\textheight]{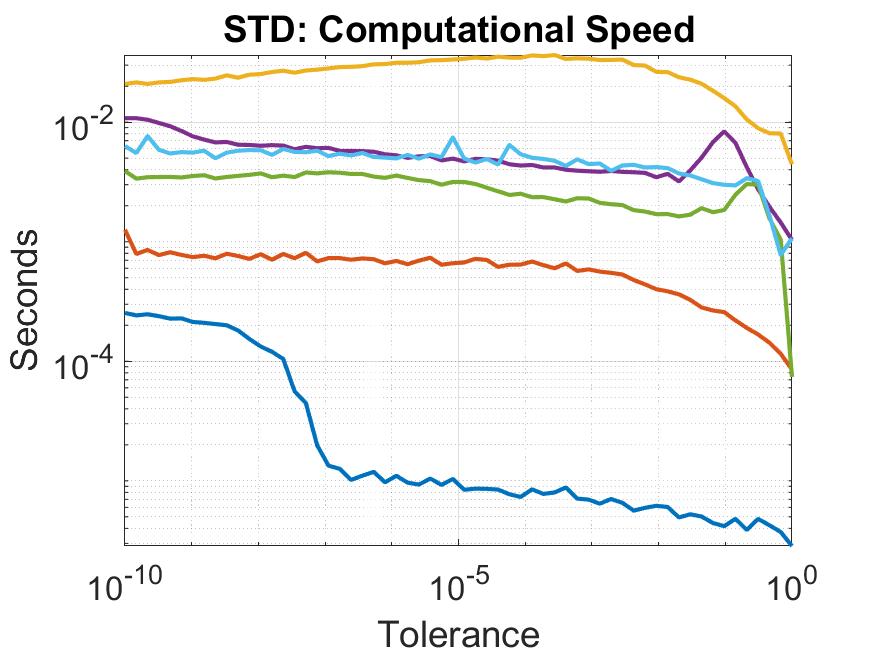}\label{fig:results_evaluation_tolerance_speed}}\hfill
  \subfloat[\it Number of iterations, {\tt Noise Level} of $10^{-1}$.]{
  \includegraphics[height=0.14\textheight]{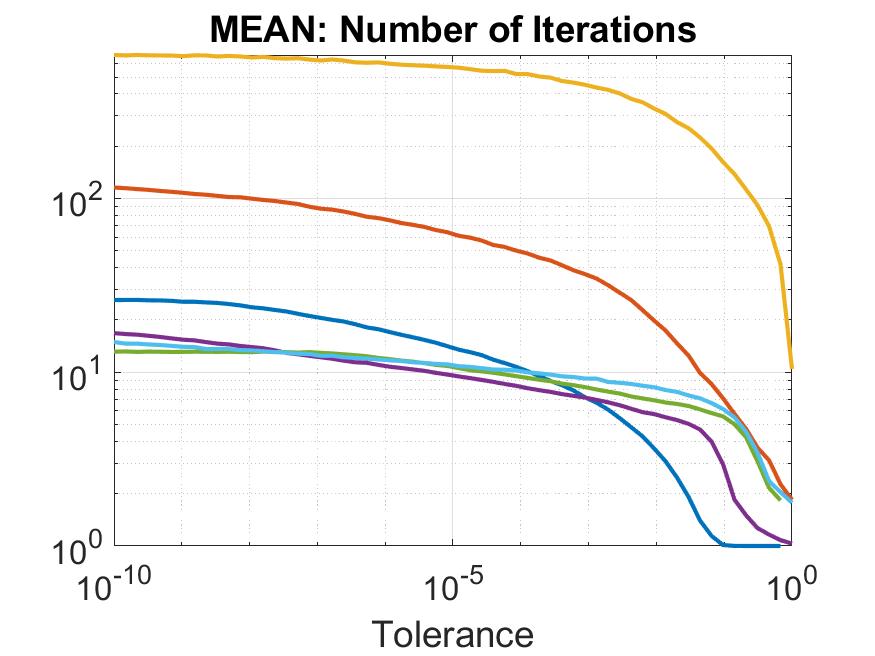}\nolinebreak
  \includegraphics[height=0.14\textheight]{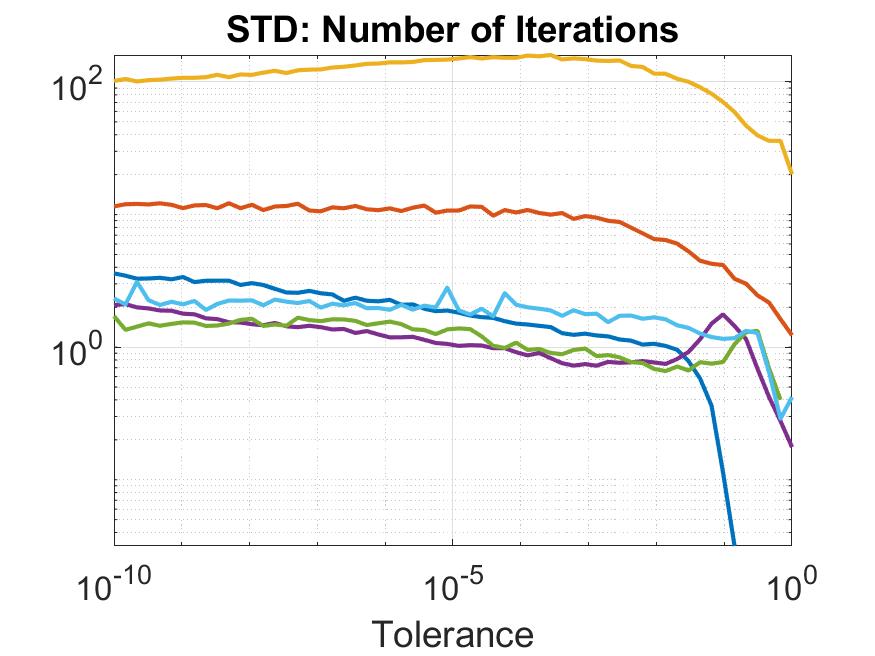}\label{fig:results_evaluation_tolerance_iter}}\\
  \subfloat[\it Computational speed, {\tt Noise Level} of $5.10^{-1}$.]{
  \includegraphics[height=0.14\textheight]{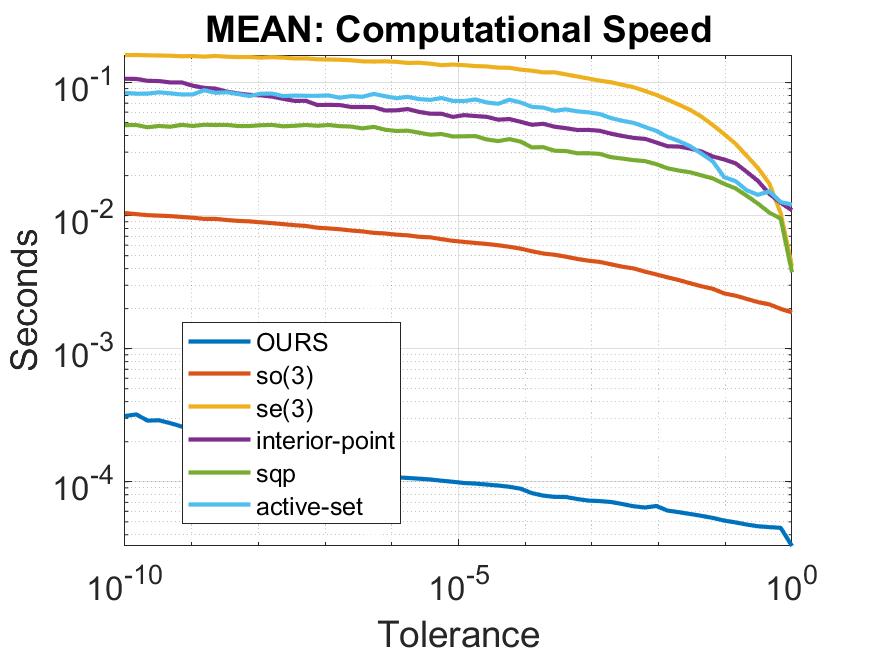}\nolinebreak
  \includegraphics[height=0.14\textheight]{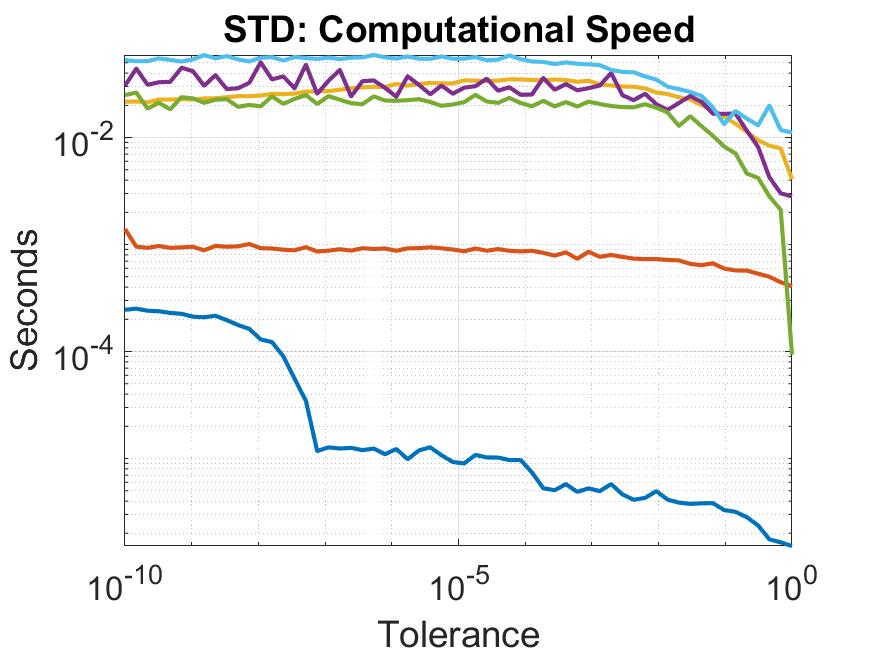}\label{fig:results_evaluation_tolerance_speed_2}} \hfill
  \subfloat[\it Number of iterations, {\tt Noise Level} of $5.10^{-1}$.]{
  \includegraphics[height=0.14\textheight]{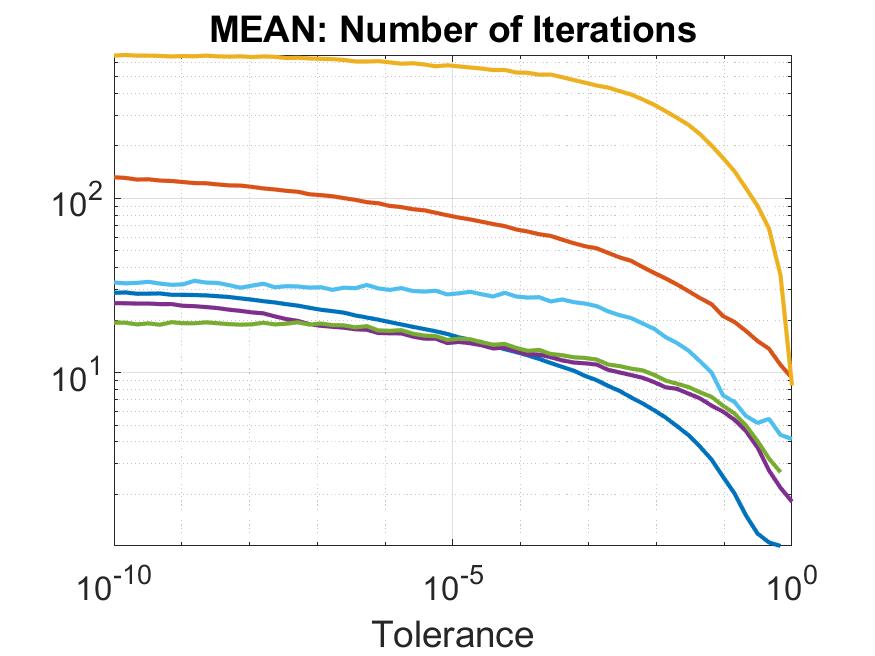}\nolinebreak
  \includegraphics[height=0.14\textheight]{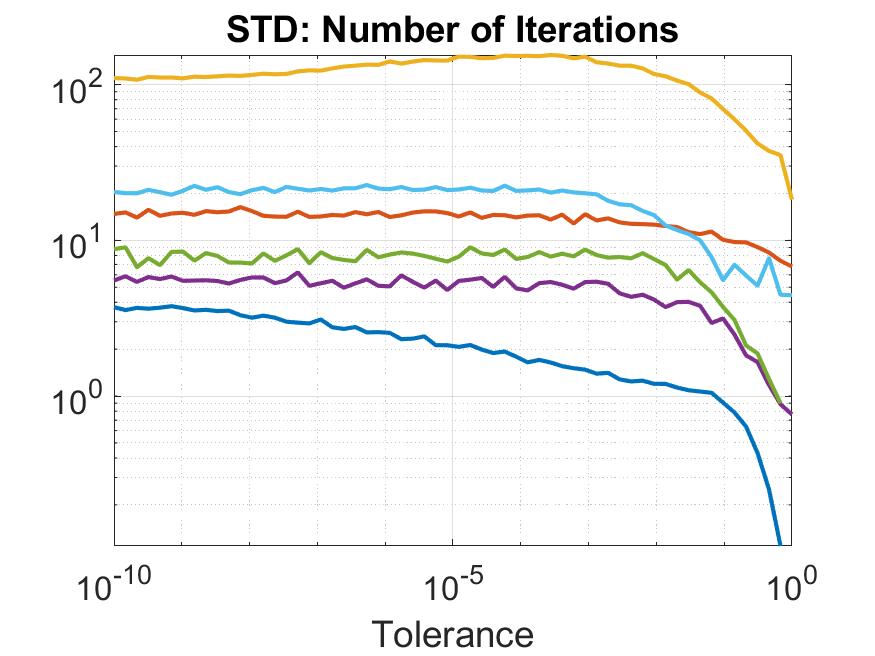}\label{fig:results_evaluation_tolerance_iter_2}}\\
  \caption{\it Comparison between Algorithm \ref{alg:ours} and the other methods, as a function of the {\tt Tolerance} considered in the algorithms. The colors of the curves and associated algorithms are identified in Fig.~\ref{fig:results_evaluation_noise}. We have considered two distinct values to the {\tt Noise Level}: $10^{-1}$ and $5.10^{-1}$, which causes the worst results for {\tt OUR} method in terms of number of iterations but not in computational speed (see Fig.~\ref{fig:results_evaluation_noise}). We evaluate the methods for both the number of iterations and computational speed, Figs.~\protect\subref{fig:results_evaluation_tolerance_iter} \& \protect\subref{fig:results_evaluation_tolerance_iter_2} and \protect\subref{fig:results_evaluation_tolerance_speed} \& \protect\subref{fig:results_evaluation_tolerance_speed_2} respectively.}
  \label{fig:results_evaluation_tolerance}
\end{figure*}

\subsection{Experimental Results}
The method described in Algorithm \ref{alg:ours} (here denoted as {\tt OUR}) is tested against the following general optimization techniques:
\begin{description}
  
  \item [{\tt so(3):}] Method based on solving the unconstrained problem in \eqref{sec:problem_reform_naive}, by the {\it Levenberg-Marquardt} algorithm;
  \item [{\tt se(3):}] Method based on solving \eqref{eq:trivial_prob_def_Direc},
  by the {\it Levenberg-Marquardt} algorithm;
  \item [{\tt interior-point:}] Solution of \eqref{eq:trivial_prob_def}, by the interior-point method \cite{Nesterov};
  \item [{\tt sqp:}] Solution of \eqref{eq:trivial_prob_def} by the sequential quadratic programming method \cite{Boggs}; and
  \item [{\tt active-set:}] Solution of \eqref{eq:trivial_prob_def} by the active-set method \cite{Murty}.
\end{description}
All the algorithms, including {\tt OUR}, were optimized (part of them have been implemented in {\tt C/C++}) and can be accessed from MATLAB. All the results shown below were implemented in this framework.

For the data-set, we first generate random rotation matrices $\mathbf{R}\in\mathcal{SO}(3)$ and random translation vectors $\mathbf{t}\in\mathbb{R}^3$. With these rotation and translation elements, we build a generalized essential matrix $\bm{ \mathcal E}\in\mathcal{X}$, as defined in \eqref{eq:generalizes_essential_matrix}.

To carry out the experiments, we propose a variation of the deviation of a generic matrix in $\mathbb{R}^{6\times 6}$ from a true generalized essential matrix. The procedure is as follows: we first generate an error matrix $\bm{\Omega}\in\mathbb{R}^{6\times 6}$, in which the respective elements are randomly generated from a normal distribution with standard deviation equal to the variable {\tt Noise Level}, and then compute the ``noisy'' matrix as $\mathbf{A} = \bm{\mathcal{E}} + \bm{\Omega}$. All the methods mentioned above are then applied.

Two tolerance values for the algorithms were selected, $10^{-9}$ \& $10^{-6}$, and we change the variable {\tt Noise Level} from $10^{-3}$ to $10^1$. Results for both the computational speed and the number of iteration are displayed in Figs.~\ref{fig:results_evaluation_noise}\subref{fig:results_evaluation_noise_speed_e9} \& \ref{fig:results_evaluation_noise}\subref{fig:results_evaluation_noise_speed_e6} and~\ref{fig:results_evaluation_noise}\subref{fig:results_evaluation_noise_iter_e9} \& \ref{fig:results_evaluation_noise}\subref{fig:results_evaluation_noise_iter_e6}, respectively. For each value of the {\tt Noise Level}, $10^3$ random trials were generated.

In addition to the evaluation of the deviation from the generalized essential matrix constraints, we have tested the proposed method against the others as a function of the tolerance of the algorithms as well (here denoted as {\tt Tolerance} value). For that purpose, we fixed a {\tt Noise Level} equal to $10^{-1}$ and to $5.10^{-1}$~\footnote{These values for the {\tt Noise Level} have been chosen to illustrate the case where our method performed worst than the other methods, with respect to the number of iterations.}, and select a {\tt Tolerance} value ranging from $10^{-15}$ to $1$. The results for both the computational speed and the number of iterations are shown in Figs.~\ref{fig:results_evaluation_tolerance}\subref{fig:results_evaluation_tolerance_speed} \& \ref{fig:results_evaluation_tolerance}\subref{fig:results_evaluation_tolerance_speed_2} and~\ref{fig:results_evaluation_tolerance}\subref{fig:results_evaluation_tolerance_iter} \& \ref{fig:results_evaluation_tolerance}\subref{fig:results_evaluation_tolerance_iter_2}, respectively. Likewise the previous case, for each level of tolerance, $10^3$ random trials were generated.

Next, we discuss these experimental results.

\subsection{Discussion}

As observed in Figs.~\ref{fig:results_evaluation_noise}\subref{fig:results_evaluation_noise_speed_e9} and \ref{fig:results_evaluation_noise}\subref{fig:results_evaluation_noise_speed_e6}, our method is significantly faster than the others. Its computational speed rarely changes as a function of the deviation from the generalized essential matrix constraints. In fact, as shown in Figs.~\ref{fig:results_evaluation_tolerance}\subref{fig:results_evaluation_tolerance_speed} and \ref{fig:results_evaluation_tolerance}\subref{fig:results_evaluation_tolerance_speed_2}, which represents the worst scenario for {\tt OUR} method, it can be seen that it performs $10^3$ times faster than any other method with the exception of {\tt so(3)}, that uses our reformulation of the problem. While our method requires around $10^{-4}$ seconds, the other methods can take up to $10^{-3}$ seconds. The experiments using the {\tt so(3)} method show a similar behaviour when compared with {\tt OUR} (almost horizontal line), around $10^1$ faster than the remaining methods. This shows that the reformulation of the problem proposed in Sec.~\ref{sec:reformulation_problem} produces better results. This is a remarkable advantage of {\tt OUR} method for applications requiring real-time computations, such as the camera relative pose estimation, which will be addressed later in Sec.~\ref{sec:real_applications_experiments}.
From these figures, we can also conclude that the use of Lie algebra techniques in {\tt se(3)} does not improve the results when compared to standard constrained optimization techniques.

In Figs.~\ref{fig:results_evaluation_noise}\subref{fig:results_evaluation_noise_iter_e9} and~\ref{fig:results_evaluation_noise}\subref{fig:results_evaluation_noise_iter_e6} one can observe that, contrarily to the general optimization techniques {\tt interior-point}, {\tt sqp}, and {\tt active-set}, the relationship between the number of iterations and {\tt Noise Level} is nearly linear for {\tt OUR}, {\tt so(3)}, and {\tt se(3)}. With the exception of the {\tt sqp} method, for low levels of noise, in general {\tt OUR} method requires less/similar number of iterations than any other technique. However, as described in the previous paragraph, the computational speed of {\tt OUR} is significantly lower for any {\tt Noise Level}, independently of the number of the required iterations. As pointed out in the previous sections, the use of {\it Lie} algebra techniques to represent rotations and transformations (methods {\tt so(3)} and {\tt se(3)}, respectively) involves a significantly larger number of iterations to reach convergence.

In addition to the evaluation in terms of deviation from the true generalized essential matrices, we have also compared all the methods with respect to the the tolerance. We have considered a {\tt Noise Level} of $10^{-1}$ and $5.10^{-1}$, which causes the worst results for {\tt OUR} method in terms of number of iterations but not in computational speed. These results are shown in Fig.~\ref{fig:results_evaluation_tolerance}. In Figs.~\ref{fig:results_evaluation_tolerance}\subref{fig:results_evaluation_tolerance_speed} and~\ref{fig:results_evaluation_tolerance}\subref{fig:results_evaluation_tolerance_speed_2}, we can observe that {\tt OUR}  is the fastest method and that {\tt so(3)} also performs well. In contrast, the method {\tt se(3)} gives the worst results.

Still about the evaluation in terms of {\tt Tolerance}, Figs.~ \ref{fig:results_evaluation_tolerance}\subref{fig:results_evaluation_tolerance_iter} and~\ref{fig:results_evaluation_tolerance}\subref{fig:results_evaluation_tolerance_iter_2} show that the number of iterations varies in a similar fashion for all the methods. However, while {\tt OUR}, {\tt interior-point}, {\tt sqp}, and {\tt active-set} techniques perform similarly in terms of number of iterations, the {\tt so(3)} and {\tt se(3)} require much more iterations. This also evidences that the use of {\it Lie} algebra techniques to get unconstrained optimization problems may simplify the problem itself, but requires in general more iterations and more running time.

\section{Results in real applications}
\label{sec:real_applications_experiments}
This section includes practical examples illustrating how Algorithm \ref{alg:ours} can be combined with other techniques to improve the results.

In Sec.~\ref{sec:advantadges} more advantages of using generalized essential matrix approximations are evidenced, in a relative pose problem for general catadioptric cameras. In Sec.~\ref{sec:validation} another application of Algorithm \ref{alg:ours}, using real data, will be shown. To conclude this section, in Sec.~\ref{sec:discussion}, we discuss the experimental results.

\subsection{A Relative Pose Problem with Non-Central Catadioptric Cameras}
\label{sec:advantadges}
Let us consider a relative position estimation problem, using a non-central catadioptric camera, composed with a perspective camera and a spherical mirror \cite{Swaminathan,Agrawal}.

\begin{figure}[t]
  \centering
  \subfloat[\it Example of the simulated world scene.]{\includegraphics[width=0.23\textwidth]{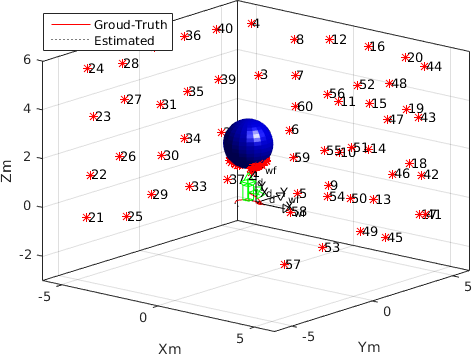}\label{fig:synthetic_world}}\hfill
  \subfloat[\it Example of the correspondence between image points.]{\includegraphics[width=0.23\textwidth]{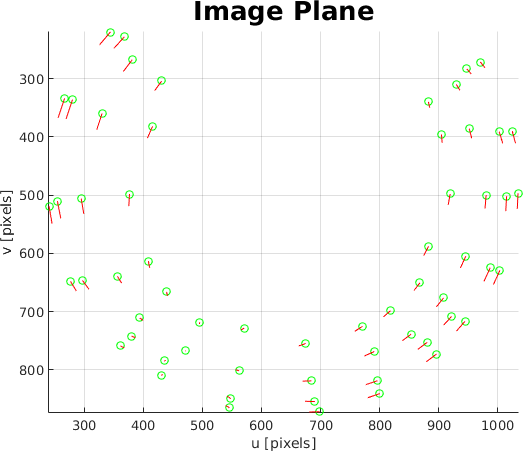}\label{fig:synthetic_image}}
\caption{\it Representation of the simulated environment created for the evaluation. At Fig.~\protect\subref{fig:synthetic_world} we show the 3D scene simulated, including: the set of 3D points; the camera system; and the subset of the path that the camera must follow. Fig~\protect\subref{fig:synthetic_image} shows an example of the projection of the 3D points in one image frame, and its correspondence with the projected points in the previous frame.}
  \label{fig:symb_environment}
\end{figure}

We synthetically generate a set of 3D points in the world (see Fig.~\ref{fig:symb_environment}\subref{fig:synthetic_world}) and, then, define a path for the camera. While the camera is following the path, we compute the projection of the 3D points onto the image of the catadioptric camera system \cite{Agrawal} (see Fig.~\ref{fig:symb_environment}\subref{fig:synthetic_image}). Then, with the knowledge of the matching between pixels at consecutive image frames, we aim at computing the rotation and translation parameters, ensuring the intersection of the respective inverse projection lines resulting from the images of the 3D points in consecutive image frames in the world.

A general technique to handle this problem can be mathematically formulated as ($\mathbf{X}$ is a matrix in $\mathbb{R}^{6\times 6}$):
\begin{equation}
  \begin{aligned}
    & \operatornamewithlimits{argmin}\limits_{\mathbf{X}}
    & & h\left(\mathbf{X}\right) \\
    & \text{subject to}
    & & \mathbf{X}_{(1:3,4:6)}^T \mathbf{X}_{(1:3,4:6)} = \mathbf{I} \\
    & & & \mathbf{X}_{(4:6,1:3)} - \mathbf{X}_{(1:3,4:6)} = \mathbf{0} \\
    & & & \mathbf{X}_{(4:6,4:6)} = \mathbf{0} \\
    & & & \mathbf{X}_{(1:3,1:3)} \mathbf{X}_{(1:3,4:6)}^T + \mathbf{X}_{(1:3,4:6)} \mathbf{X}_{(1:3,1:3)}^T  = \mathbf{0} \\
  \end{aligned},
  \label{eq:trivial_relative_pose_problem}
\end{equation}
where
\begin{equation}
  h\left(\mathbf{X}\right) = \sum_{i=0}^{N}{\left.\mathbf{l}_i^{\left(\mathcal{L}\right)}\right.^T \mathbf{X}\ \mathbf{l}_i^{\left(\mathcal{R}\right)}},
\end{equation}
$\mathbf{l}_i^{\left(\mathcal{L}\right)}$ and $\mathbf{l}_i^{\left(\mathcal{R}\right)}$ represent the matching between the image projection lines on left and right cameras, respectively, and $N$ is the number of matching points.

We consider six distinct methods for the computation of the relative pose, based on the estimation of the generalized essential matrix:
\begin{description}
  \item [{\tt Full}] Denotes the relative pose estimation that aligns 3D straight lines in the world to ensure that they intersect, by the optimization problem \eqref{eq:trivial_relative_pose_problem}. A tolerance of $10^{-9}$ was considered for the constraints;
  \item [{\tt Without Constraints}] Denotes a method similar to {\tt Full}, i.e. the problem of \eqref{eq:trivial_relative_pose_problem}. However, in this case, a different value of $10^{-1}$ was considered for the tolerance of constraints;
\item [{\tt OUR + WC:}] Consists in, first, estimating an initial solution using the {\tt Without Constraints} method and, then, applying {\tt OUR} method to estimate a true generalized essential matrix (Algorithm \ref{alg:ours}), with tolerance $10^{-9}$ for the constraints;
  \item [{\tt interior-point + WC:}] Same as {\tt OUR + WC} but now the approximation is given by solving \eqref{eq:trivial_prob_def} with the {\tt interior-point};
  \item [{\tt SQP + WC:}] Same as {\tt interior-point + WC}, but with the approximation of \eqref{eq:trivial_prob_def} obtained with the {\tt sqp} algorithm;
  \item [{\tt active-set + WC:}] Same as {\tt interior-point + WC} but now \eqref{eq:trivial_prob_def} is  solved by {\tt active-set}.
\end{description}

\begin{figure}
  \centering
  \includegraphics[width=.5\textwidth]{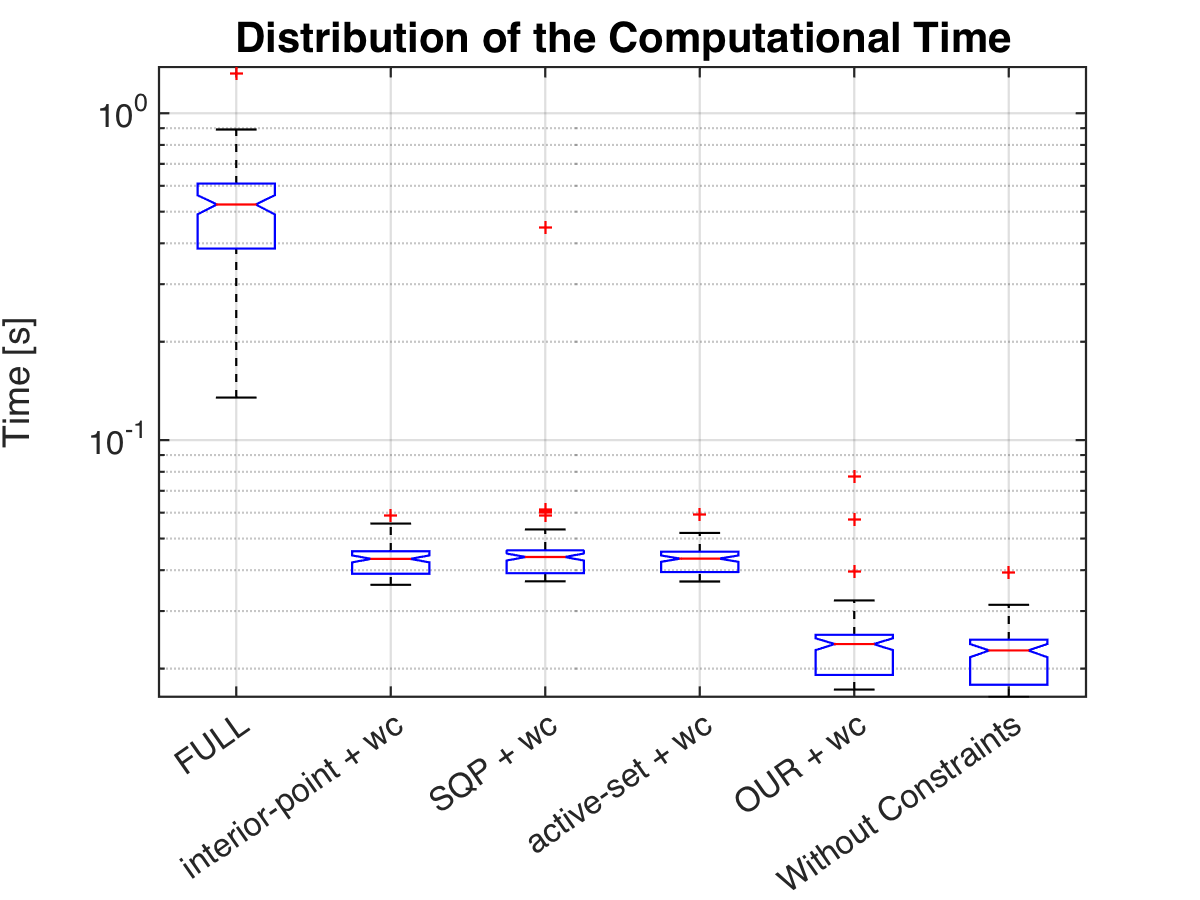}
  \caption{\it Results for the distribution of the computational time required for computing the camera relative pose, for each of the methods described in Sec.~\ref{sec:advantadges}.}
  \label{fig:evaluation}
\end{figure}

The results for the distribution of the computational time required to compute each image frame are shown in Fig.~\ref{fig:evaluation} (box plot graph). These results are commented later in Sec.~\ref{sec:discussion}.

Note that now we are dealing with a different optimization problem from \eqref{eq:trivial_prob_def}, despite the constraints coincide.

\subsection{Experiments with Real Data}
\label{sec:validation}
To conclude the experimental results, we apply Algorithm~\ref{alg:ours} to an absolute pose estimation problem in the framework of general camera models, and known coordinates of 2D straight lines in the world coordinate system \cite{Miraldo2}. We consider a non-central catadioptric camera (with a spherical mirror) and moved the camera along a path in the lab.

\begin{figure}[t]
  \centering
  \includegraphics[height=.13\textheight]{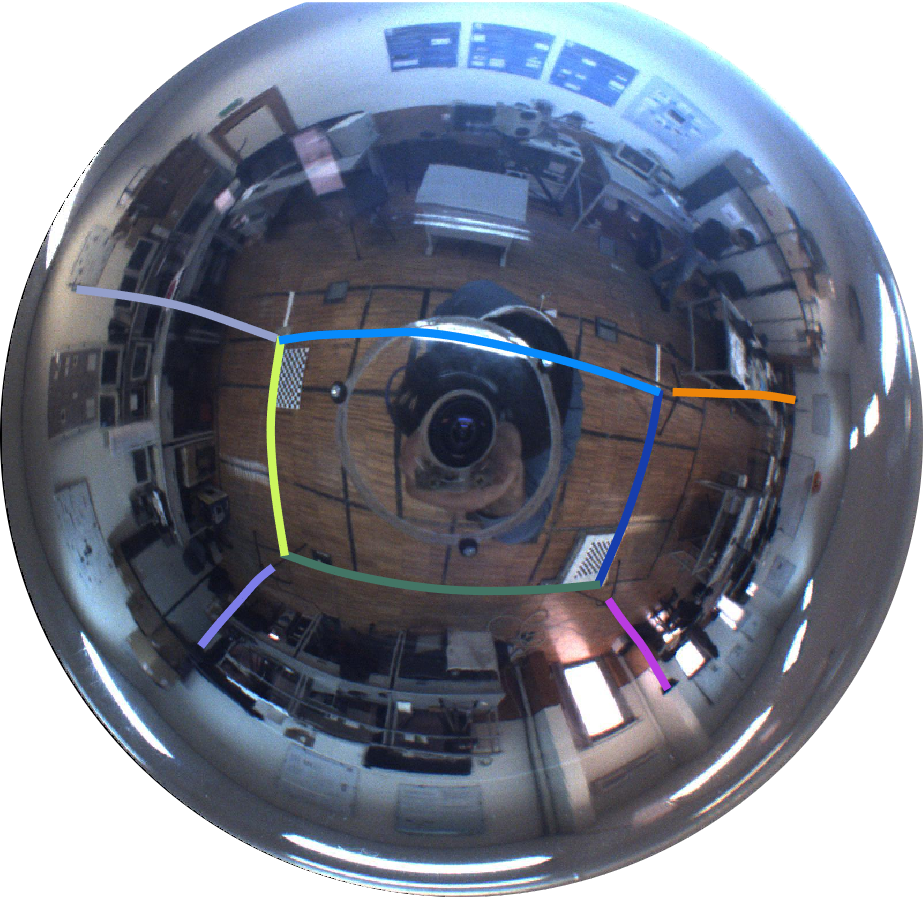} \quad
  \includegraphics[height=.13\textheight]{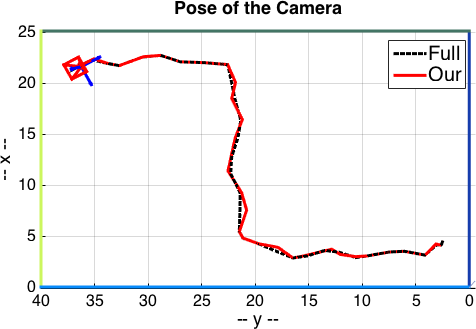}
  \caption{\it Results obtained with real images. At the left, we show an example of an image acquired by the non-central catadioptric camera. On the right, we show the results for the trajectory (top view) computed with the {\tt FULL} and {\tt OUR+WC} methods, described in Sec.~\ref{sec:advantadges}.}
  \label{fig:real_data_dataset}
\end{figure}

3D lines in the world are associated with a set of pixels in the image (see Fig.~\ref{fig:real_data_dataset}). The goal is to find the generalized essential matrix $\bm{\mathcal{E}}$ that aligns the 3D inverse projection lines from these pixels with the known 3D straight lines in the world, in order to guarantee their intersection.

This problem can be solved by the same strategy proposed in the previous subsection, i.e. by using the optimization problem of \eqref{eq:trivial_relative_pose_problem}, but in this case with the following objective function:
\begin{equation}
  h\left(\mathbf{X}\right) = \sum_{i=0}^{M}{\sum_{j=0}^{N_i}{\left.\mathbf{l}_{i}^{\left(\mathcal{W}\right)}\right.^T \mathbf{X}\ \mathbf{l}_{i,j}^{\left(\mathcal{C}\right)}}},
\end{equation}
where: $\mathbf{l}_i^{\left(\mathcal{W}\right)}$ represent the known 3D straight lines in the world; $\mathbf{l}_{i,j}^{\left(\mathcal{C}\right)}$ the inverse projection lines corresponding pixels that are images of the $i^{\text{th}}$ line; $N_i$ are the number of image pixels that are images of the line $i$; and $M$ is the number of known 3D straight lines in the world.

We consider the six methods proposed in Sec.~\ref{sec:advantadges}. A comparison between the trajectories of {\tt OUR + WC}\footnote{Notice that the other approximate techniques would give the same results as {\tt OUR + WC}. The difference between them depends only on the required computational time.} method and the {\tt FULL} are shown in Fig.~\ref{fig:real_data_dataset}. The distribution of the required computational time for each frame is shown in Fig.~\ref{fig:real_images_results}.

\subsection{Discussion of the Results}
\label{sec:discussion}

\begin{figure}[t]
  \centering
  \includegraphics[width=.50\textwidth]{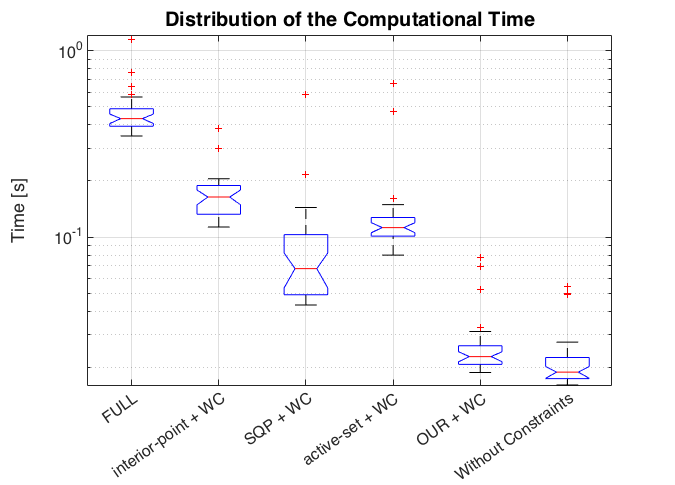}
  \caption{\it Distribution of the computational time obtained for the absolute pose problem using a real non-central catadioptric camera. We consider all the algorithms described in Sec.~\ref{sec:advantadges}.}
  \label{fig:real_images_results}
\end{figure}

In this subsection, we discuss the experimental results shown in the previous subsections. We start by analyzing the results of the approximation of general $6\times 6$ matrices, in which we compare the performance of the proposed method against the direct solution (these are the main results of this paper) in a non-central catadioptric relative camera pose. Next, we discuss the experiments with real data, in which we compare the performance of our approximation technique against the direct solution, in an absolute pose problem.

In all these tests, we have imposed $m = 100$ as the maximum number of iterations for Algorithm~\ref{alg:ours}. It is worth noting that the algorithm has never reached such a number of iterations, which means that it always converged.

Notice that one of the main contributions of this paper (Sec.~\ref{sec:efficient_solution}) is not to estimate the relative pose of an imaging device, but, instead, to find generalized essential matrices from general $6\times 6$ matrices (that do not verify \eqref{eq:trivial_prob_def}).

\vspace{.25cm}
\noindent
{\bf Evaluation in a non-central catadioptric relative pose problem: }
When considering the experiments carried out in Sec.~\ref{sec:advantadges}, first, we conclude that increasing the tolerance of constraints on the {\tt FULL} algorithm (i.e., not fully considering the underlying constraints of the generalized essential matrix \eqref{eq:trivial_relative_pose_problem}) and, then, recover a true generalized essential matrix, by Algorithm~\ref{alg:ours}, leads to significant savings in computational time. See the comparison between {\tt FULL} and all the other methods in Figs.~\ref{fig:evaluation} and~\ref{fig:real_images_results}. It can be seen that the differences between {\tt OUR + WC} and {\tt Without Constraints} (the optimization without fully considering the underlying constraints of the generalized essential matrix) can be neglected, while this does not happen for the direct solution. We recall that the {\tt Without Constraints} method does not produce a true generalized essential matrix, while the other ones do.

Besides, from Fig.~\ref{fig:symb_environment}, one can conclude that this procedure (compute $\mathbf{A}$ and then find the closest $\mathbf{X}$) does not diminish the results significantly.

To conclude, one can see that estimating $\mathbf{A}$ and, then, find $\mathbf{X}$ that approximates $\mathbf{A}$ (see \eqref{eq:problem_def}) will result in a much faster algorithm than looking directly for $\mathbf{X}$.

\vspace{.25cm}
\noindent
{\bf Validation using Real Data:}
To conclude the experimental results, we validate the proposed fitting technique (Algorithm~\ref{alg:ours}) against the direct solution (with the above mentioned general optimization techniques on the problem defined in \eqref{eq:trivial_prob_def}) in a real application of an absolute camera pose estimation, when using non-central catadioptric cameras, see Fig.~\ref{fig:real_data_dataset}.

From Fig.~\ref{fig:real_images_results}, we see that, while the use of the direct solution and all the three general optimization techniques will have an impact on the computation time (see the results for {\tt interior-point+WC}, {\tt SQP+WC}, and {\tt active-set} against {\tt Without Constraints}), the difference between {\tt OUR+WC} and {\tt Without Constraints} can be neglected, being much faster than the {\tt FULL} technique.

Still from Fig.~\ref{fig:real_data_dataset}, one can conclude that approximating $\mathbf{X}$ (a true generalized essential matrix) from a general matrix $\mathbf{A}$ does not degrade significantly the results, being much faster than estimating $\mathbf{X}$ directly (that is shown by comparing {\tt FULL} and {\tt OUR+WC} in Fig.~\ref{fig:real_images_results}).

Although these tests have different goals (one is related with the relative camera pose and the other with the absolute pose), these results are very similar to the ones presented in Sec.~\ref{sec:advantadges}, validating the results using synthetic data.
\section{Conclusions}
\label{sec:conclusions}
To the best of our knowledge, we are the first to investigate the problem of fitting generalized essential matrices, from general $6\times 6$ matrices, and its implications. We have defined the problem and proposed some general constrained optimization techniques that can be used to determine a solution. One of the issues of the general optimization techniques is the large amount of constraints involved (33 constraints). To get rid off those constraints, two unconstrained formulations of the problem have been proposed. However, they have led to unsatisfactory results. Then, we have proposed an efficient technique by optimizing on matrix manifolds. More specifically, we define a constrained optimization problem on the rotation group. A suitable efficient iterative method, of steepest descent-type, has been then developed to solve the problem, which ensures that each iteration lies on the manifold of rotation matrices. A large set of experiments has shown that such a method is the fastest.

We have also presented some results to show the advantages of using approximating techniques such as the one presented in this paper, in real applications.  We evaluate our method against the direct solution in relative and absolute pose problems (the latter with real data), in which we prove that: 1) estimating $\mathbf{A}$ (general $6\times 6$ matrix) and then fitting $\bm{\mathcal{E}}$ (a true generalized essential matrix) speed up significantly the required computational time; and 2) there is no significant deterioration of the obtained results. We also concluded that, contrarily to the direct solution, when using our method, the required additional computational time (i.e., the computation time that is required after the estimation of $\mathbf{A}$) can be neglected.

\bibliographystyle{spmpsci}      
\bibliography{files/bib.bib}   

\begin{thebibliography}{10}
\providecommand{\url}[1]{{#1}}
\providecommand{\urlprefix}{URL }
\expandafter\ifx\csname urlstyle\endcsname\relax
  \providecommand{\doi}[1]{DOI~\discretionary{}{}{}#1}\else
  \providecommand{\doi}{DOI~\discretionary{}{}{}\begingroup
  \urlstyle{rm}\Url}\fi

\bibitem{Abrudan}
Abrudan, T.E., Eriksson, J., Koivunen, V.: Steepest descent algorithms for
  optimization under unitary matrix constraint.
\newblock IEEE Trans. Signal Processing \textbf{56}(3), 635--650 (2008)

\bibitem{Absil}
Absil, P.A., Mahoney, R., Sepulchre, R.: Optimization Algorithms on Matrix
  Manifolds.
\newblock Princeton University Press, New Jersey, USA (2007)

\bibitem{Agrawal}
Agrawal, A., Taguchi, Y., Ramalingam, S.: Analytical forward projection for
  axial non-central dioptric and catadioptric cameras.
\newblock In: European Conf. Computer Vision (ECCV), pp. 129--143 (2010)

\bibitem{Boggs}
Boggs, P.T., Tolle, J.W.: Sequential quadratic programming.
\newblock Acta Numerica \textbf{4}, 1--51 (1995)

\bibitem{Campos}
Campos, J., Cardoso, J.R., Miraldo, P.: Poseamm: A unified framework for
  solving pose problems using an alternating minimization method.
\newblock In: IEEE Int'l Conf. Robotics and Automation (ICRA), pp. 3493--3499
  (2019)

\bibitem{Cardoso}
Cardoso, J.R., Kenney, C.S., Leite, F.S.: Computing the square root and
  logarithm of a real $p$-orthogonal matrix.
\newblock Applied Numerical Mathematics \textbf{46}(2), 173--196 (2003)

\bibitem{Cardoso10}
Cardoso, J.R., Leite, F.S.: Exponentials of skew-symmetric matrices and
  logarithms of orthogonal matrices.
\newblock J. Comput. Appl. Math \textbf{233}(11), 2867--2875 (2010)

\bibitem{Edelman}
Edelman, A., Arias, T.A., Smith, S.T.: The geometry of algorithms with
  orthogonal constraints.
\newblock SIAM Journal on Matrix Analysis and Applications (SIMAX)
  \textbf{20}(2), 303--353 (1998)

\bibitem{Golub}
Golub, G.H., Loan, C.F.V.: Matrix Computations, 3 edn.
\newblock Johns Hopkins University Press (1996)

\bibitem{Grossberg}
Grossberg, M.D., Nayar, S.K.: A general imaging model and a method for finding
  its parameters.
\newblock In: IEEE Int'l Conf. Computer Vision (ICCV), vol.~2, pp. 108--115
  (2001)

\bibitem{Gupta}
Gupta, R., Hartley, R.I.: Linear pushbroom cameras.
\newblock IEEE Trans. Pattern Analysis and Machine Intelligence (T-PAMI)
  \textbf{19}(9), 963--975 (1997)

\bibitem{Hartley97}
Hartley, R.: In defence of the eight-point algorithm.
\newblock IEEE Trans. Pattern Analysis and Machine Intelligence (T-PAMI)
  \textbf{19}(6), 580--593 (1997)

\bibitem{Hartley}
Hartley, R., Zisserman, A.: Multiple View Geometry in Computer Vision, 2 edn.
\newblock Cambridge University Press (2003)

\bibitem{Jiang}
Jiang, B., Dai, Y.H.: A framework of constraint preserving update schemes for
  optimization on stiefel manifold.
\newblock Mathematical Programming \textbf{153}, 535--575 (2015)

\bibitem{Kim}
Kim, J.H., Li, H., Hartley, R.: Motion estimation for nonoverlapping
  multicamera rigs: Linear algebraic and $l_{\infty}$ geometric solutions.
\newblock IEEE Trans. Pattern Analysis and Machine Intelligence (T-PAMI)
  \textbf{32}(6), 1044--1059 (2010)

\bibitem{Kneip}
Kneip, L., Li, H.: Efficient computation of relative pose for multi-camera
  systems.
\newblock In: IEEE Conf. Computer Vision and Pattern Recognition (CVPR), pp.
  446--453 (2014)

\bibitem{Lee}
Lee, G.H., Pollefeys, M., Fraundorfer, F.: Relative pose estimation for a
  multi-camera system with known vertical direction.
\newblock In: IEEE Conf. Computer Vision and Pattern Recognition (CVPR), pp.
  540--547 (2014)

\bibitem{Li}
Li, H., Hartley, R., hak Kim, J.: A linear approach to motion estimation using
  generalized camera models.
\newblock In: IEEE Conf. Computer Vision and Pattern Recognition (CVPR), pp.
  1--8 (2008)

\bibitem{Luenberger}
Luenberger, D.G., Ye, Y.: Linear and Nonlinear Programming, 3 edn.
\newblock Springer: New York (2008)

\bibitem{Lutkepohl}
Lutkepohl, H.: Handbook of Matrices.
\newblock Jonh Wiley and Sons (1996)

\bibitem{Ma}
Ma, Y., Soatto, S., Kosecka, J., Sastry, S.S.: An Invitation to 3-D Vision:
  From Images to Geometric Models.
\newblock SpringerVerlag (2003)

\bibitem{Manton}
Manton, J.H.: Optimization algorithms exploiting unitary constraints.
\newblock IEEE Trans. Signal Processing \textbf{50}(3), 635--650 (2002)

\bibitem{Micusik}
Micusik, B., Pajdla, T.: Autocalibration \& 3d reconstruction with non-central
  catadioptric cameras.
\newblock In: IEEE Conf. Computer Vision and Pattern Recognition (CVPR), pp.
  58--65 (2004)

\bibitem{Miraldo3}
Miraldo, P., Araujo, H.: Calibration of smooth camera models.
\newblock IEEE Trans. Pattern Analysis and Machine Intelligence (T-PAMI)
  \textbf{35}(9), 2091--2103 (2013)

\bibitem{Miraldo}
Miraldo, P., Araujo, H.: Generalized essential matrix: Properties of the
  singular value decomposition.
\newblock Image and Vision Computing (IVC) \textbf{34}, 45--50 (2015)

\bibitem{Miraldo2}
Miraldo, P., Araujo, H., Goncalves, N.: Pose estimation for general cameras
  using lines.
\newblock IEEE Trans. Cybernetics \textbf{45}(10), 2156--2164 (2015)

\bibitem{Moler}
Moler, C., Loan, C.V.: Nineteen dubious ways to compute the exponential of a
  matrix, twenty-five years later.
\newblock SIAM Review \textbf{45}(1), 3--49 (2003)

\bibitem{Mouragnon}
Mouragnon, E., Lhuillier, M., Dhome, M., Dekeyser, F., Sayd, P.: Generic and
  real-time structure from motion using local bundle adjustment.
\newblock Image and Vision Computing (IVC) \textbf{27}, 1178--1193 (2009)

\bibitem{Murray}
Murray, R.M., Li, Z., Sastry, S.S.: A Mathematical Introduction to Robotic
  Manipulation.
\newblock CRC Press (1994)

\bibitem{Murty}
Murty, K.G., Yu, F.T.: Linear Complementarity, Linear and Nonlinear
  Programming.
\newblock Berlin: Heldermann (1988)

\bibitem{Nalwa}
Nalwa, V.S.: A true omnidirectional viewer.
\newblock Tech. rep., Bell Laboratories (1995)

\bibitem{Nayar}
Nayar, S.K.: Catadioptric omnidirectional camera.
\newblock In: IEEE Conf. Computer Vision and Pattern Recognition (CVPR), pp.
  482--488 (2004)

\bibitem{Nesterov}
Nesterov, Y., Nemirovskii, A.: Interior-Point Polynomial Algorithms in Convex
  Programming.
\newblock SIAM: Studies in Applied and Numerical Mathematics (1994)

\bibitem{Nister}
Nister, D., Naroditsky, O., Bergen, J.: Visual odometry.
\newblock In: IEEE Conf. Computer Vision and Pattern Recognition (CVPR),
  vol.~2, pp. 652--659 (2004)

\bibitem{Nocedal}
Nocedal, J., Wright, S.: Numerical Optimization.
\newblock Springer-Verlag: New York (1999)

\bibitem{Pless}
Pless, R.: Using many cameras as one.
\newblock In: IEEE Conf. Computer Vision and Pattern Recognition (CVPR),
  vol.~2, pp. 587--593 (2003)

\bibitem{Pottmann}
Pottmann, H., Wallner, J.: Computational Line Geometry.
\newblock Springer-Verlag Berlin Heidelberg (2001)

\bibitem{Ramalingam}
Ramalingam, S., Lodha, S.K., Sturm, P.: A generic structure-from-motion
  framework.
\newblock Computer Vision and Image Understanding (CVIU) \textbf{103}, 218--228
  (2006)

\bibitem{Stewenius}
Stewenius, H., Oskarsson, M., Astrom, K., Nister, D.: Solutions to minimal
  generalized relative pose problems.
\newblock In: Workshop on Omnidirectional Vision (2005)

\bibitem{Sturm}
Sturm, P.: Multi-view geometry for general camera models.
\newblock In: IEEE Conf. Computer Vision and Pattern Recognition (CVPR),
  vol.~1, pp. 206--212 (2005)

\bibitem{Sturm2}
Sturm, P., Ramalingam, S.: A generic concept for camera calibration.
\newblock In: European Conf. Computer Vision (ECCV), pp. 1--13 (2004)

\bibitem{Swaminathan}
Swaminathan, R., Grossberg, M.D., Nayar, S.K.: Non-single viewpoint
  catadioptric cameras: Geometry and analysis.
\newblock Int'l J. Computer Vision (IJCV) \textbf{66}(3), 211--229 (2002)

\bibitem{Treibitz}
Treibitz, T., Schechner, Y.Y., Singh, H.: Flat refractive geometry.
\newblock In: IEEE Conf. Computer Vision and Pattern Recognition (CVPR), pp.
  1--8 (2008)

\bibitem{Wen}
Wen, Z., Yin, W.: A feasible method for optimization with orthogonality
  constraints.
\newblock Mathematical Programming \textbf{142}, 397--434 (2013)

\bibitem{Yamaguchi}
Yamaguchi, K., McAllester, D., Urtasun, R.: Robust monocular epipolar flow
  estimation.
\newblock In: IEEE Conf. Computer Vision and Pattern Recognition (CVPR), pp.
  1863--1869 (2013)

\bibitem{Zaragoza}
Zaragoza, J., Chin, T.J., Tran, Q.H., Brown, M.S., Suter, D.:
  As-projective-as-possible image stitching with moving dlt.
\newblock In: IEEE Conf. Computer Vision and Pattern Recognition (CVPR), pp.
  2339--2344 (2013)

\end{thebibliography}

\vfill

\end{document}